%% file: main.tex
\documentclass[11pt]{style}

\usepackage{parskip}
\usepackage[numbers]{natbib}
\usepackage{lmodern}
\usepackage{subcaption}
\usepackage{xspace}
\usepackage{tikz}
\usepackage{hyperref}
\usepackage{url}
\usepackage{booktabs}
\usepackage{listings}
\usepackage{tcolorbox}
\tcbuselibrary{breakable,skins,listings}
\usepackage{tabularx}
\usepackage{array}
\usepackage{rotating}
\usepackage{ulem}
\usepackage{float}
\usepackage{wrapfig}
\usepackage{makecell}
\usepackage{pgfplots}
\usepackage{bbm}
\usepackage{placeins}
\pgfplotsset{compat=1.18}
\usepackage[nameinlink,capitalise]{cleveref}

\theoremstyle{remark}

\newcommand{\titleLogo}{%
    $\raisebox{-1.4mm}
    {\includegraphics[width=0.05\textwidth]{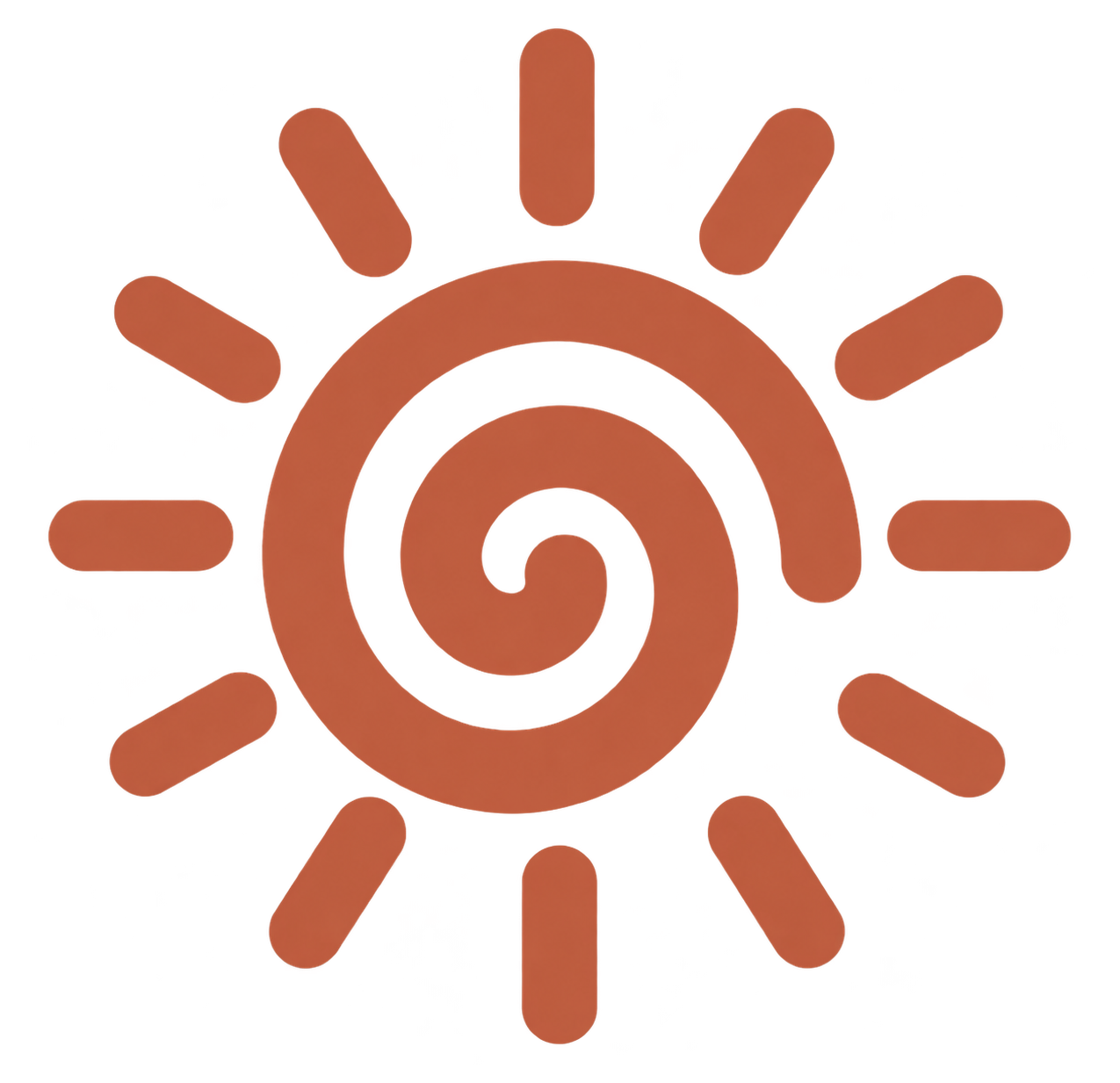}}$ 
}

\makeatletter
\patchcmd{\archer@maketitle@onecolumn}{\vskip11pt}{\vskip3pt}{}{}
\makeatother

\title{\titleLogo \benchmarkname: Evaluating Satisfactory Completion of Live User-Facing Digital Work}

\runningtitle{%
  \benchmarkname: Evaluating Satisfactory Completion of Live User-Facing Digital Work
}

\author[1,\textbf{*}]{Fangrui Huang}
\author[2,\textbf{*}]{Bingyang Wang}
\author[3]{Chuhan Li}
\author[4]{Yijiang Li}
\author[1]{Arpandeep Khatua}

\author[5]{Austin Shen}
\author[6]{Riki R Z Choi}
\author[1]{Cuiyuanxiu Chen}
\author[1]{Jin Huang}
\author[1]{Zachary Robertson}

\author[9]{Kate Lin}
\author[1]{Akshay Paruchuri}
\author[8]{Sheng Wang} 
\author[7]{Sijun Tan}
\author[9]{\authorcr Xin Liu} 
\author[10]{ Zhengmin Ni}
\author[1]{Sanmi Koyejo}
\author[1]{Ehsan Adeli}

\affil[1]{Stanford University}
\affil[2]{Georgia Institute of Technology}
\affil[3]{University of California, Santa Barbara}
\makeatletter
\renewcommand{\AB@affilsep}{\protect\\\protect\Affilfont}
\makeatother
\affil[4]{University of California, San Diego}
\affil[5]{University of Michigan}
\affil[6]{Boston University}
\affil[7]{University of California, Berkeley}
\affil[8]{The University of Hong Kong}
\affil[9]{Google Research}
\affil[10]{flowith AI}

\begingroup

\footnotetext[1]{Equal contribution.}
\endgroup

\begin{document}
\input{premeable}
\maketitle
\input{sections/0-abstract}

\input{sections/1-introduction}

\input{sections/2-related_work}
\input{sections/3-benchmark_construction}
\input{sections/4-experiments}
\input{sections/5-main_results}

\input{sections/6-conclusion}

\clearpage
\bibliographystyle{abbrv}
\bibliography{references}
\clearpage
\appendix
\input{sections/appendix}

\end{document}

%% file: premeable.tex
\newif\ifcomments
\commentstrue

\newcommand{\authorcomment}[3]{%
  \ifcomments
    \textcolor{#1}{\bfseries\small [#2: #3]}%
  \fi
}

\newcommand{\authornew}[2]{%
  \ifcomments
    \textcolor{#1}{#2}%
  \else
    #2%
  \fi
}

\definecolor{darkPurple}{HTML}{534d6b} 
\definecolor{sageGreen}{HTML}{5B9877} 

\newcommand{\bcmark}{\color{darkPurple}{\ding{51}}}%
\newcommand{\cmark}{\color{sageGreen}{\ding{51}}}%
\newcommand{\xmark}{\color{Maroon!70}{\ding{55}}}%

\newcommand{\notcheckmark}{\textcolor{black}{\bcmark\kern-1.1ex\raisebox{.7ex}{\rotatebox[origin=c]{125}{--}}}\color{black}}

\newcommand{\cl}[1]{\authorcomment{purple}{Chuhan}{#1}}
\newcommand{\clnew}[1]{\authornew{purple}{#1}}

\newcommand{\fr}[1]{\authorcomment{blue}{Fangrui}{#1}}
\newcommand{\frnew}[1]{\authornew{blue}{#1}}

\newcommand{\benchmarkname}{\textsc{EverydayBench}\xspace}

%% file: sections/0-abstract.tex
\begin{abstract}

Everyday digital tasks, from selecting products and planning trips to conducting research and creating artifacts, require AI agents to deliver useful outcomes rather than merely retrieve web information.
Yet evaluating these outcomes on the live web is challenging, as a single request may admit multiple valid results, and changing web information can alter which results remain valid over time.
Although existing benchmarks address aspects of this setting, they do not fully integrate predefined success criteria, current evidence verification, and deliverable-quality assessment for agent-selected outcomes.
To address this gap, we introduce \benchmarkname, a benchmark of 77 human-authored tasks spanning eight domains of everyday life and work.
Each task specifies success criteria before execution without prescribing an answer or workflow, allowing agents to select their own outcomes.
To evaluate submissions under fixed success criteria, \benchmarkname combines (1) deterministic checks of explicit constraints, (2) verification of claims against agent-cited webpages retrieved at evaluation time, and (3) rubric-based assessment of final deliverable quality.
We further validate the protocol through human judgments, repeated scoring under fixed evidence, and controlled grounding failures.
Across 23 reported agent--model configurations, Claude Opus 5.5 and Claude Fable 5.1 tie for the highest strict pass rate at 54.55\%; a controlled GPT-6 Sol effort sweep ranges from 29.87\% at low effort to 45.45\% at max effort.
Reliability varies substantially by task type: GPT-5.6 Sol leads lookup, Fable 5.1 leads planning, Opus 5.5 and Fable 5.1 tie on comparison, and GPT-6 Astra and Sonnet 5.5 tie on customized deliverables.
Together, these findings highlight the difficulty of delivering complete, well-supported, and usable outcomes for everyday user requests.

\par\medskip
\noindent
\begin{minipage}[b]{0.72\linewidth}
\textbf{Project Website:}
\href{https://everydaybench.stanford.edu}{https://everydaybench.stanford.edu}\\
\textbf{Corresponding authors:}
Fangrui Huang (\href{mailto:fangruih@stanford.edu}{fangruih@stanford.edu}), 
Ehsan Adeli (\href{mailto:eadeli@stanford.edu}{ eadeli@stanford.edu})

\end{minipage}%
\hfill
\begin{minipage}[b]{0.26\linewidth}
\raggedleft
\includegraphics[height=1.0cm]{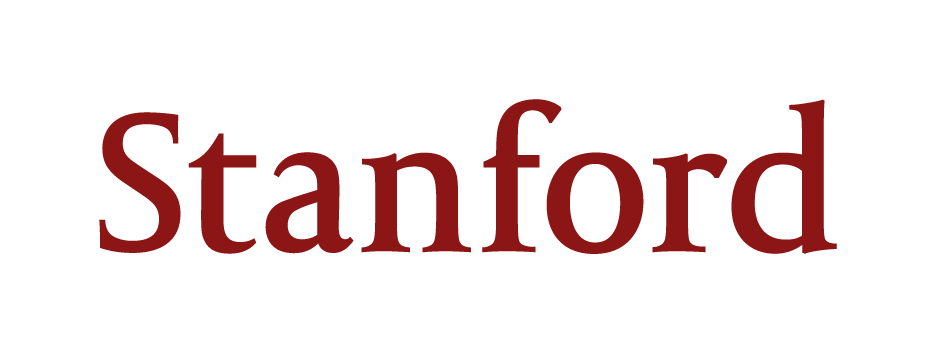}%
\hspace{0.8em}%

\end{minipage}

\end{abstract}

\input{figure_tex/fig_overview}

%% file: figure_tex/fig_overview.tex
\begin{figure}[h]
    \centering
    \includegraphics[width=\textwidth]{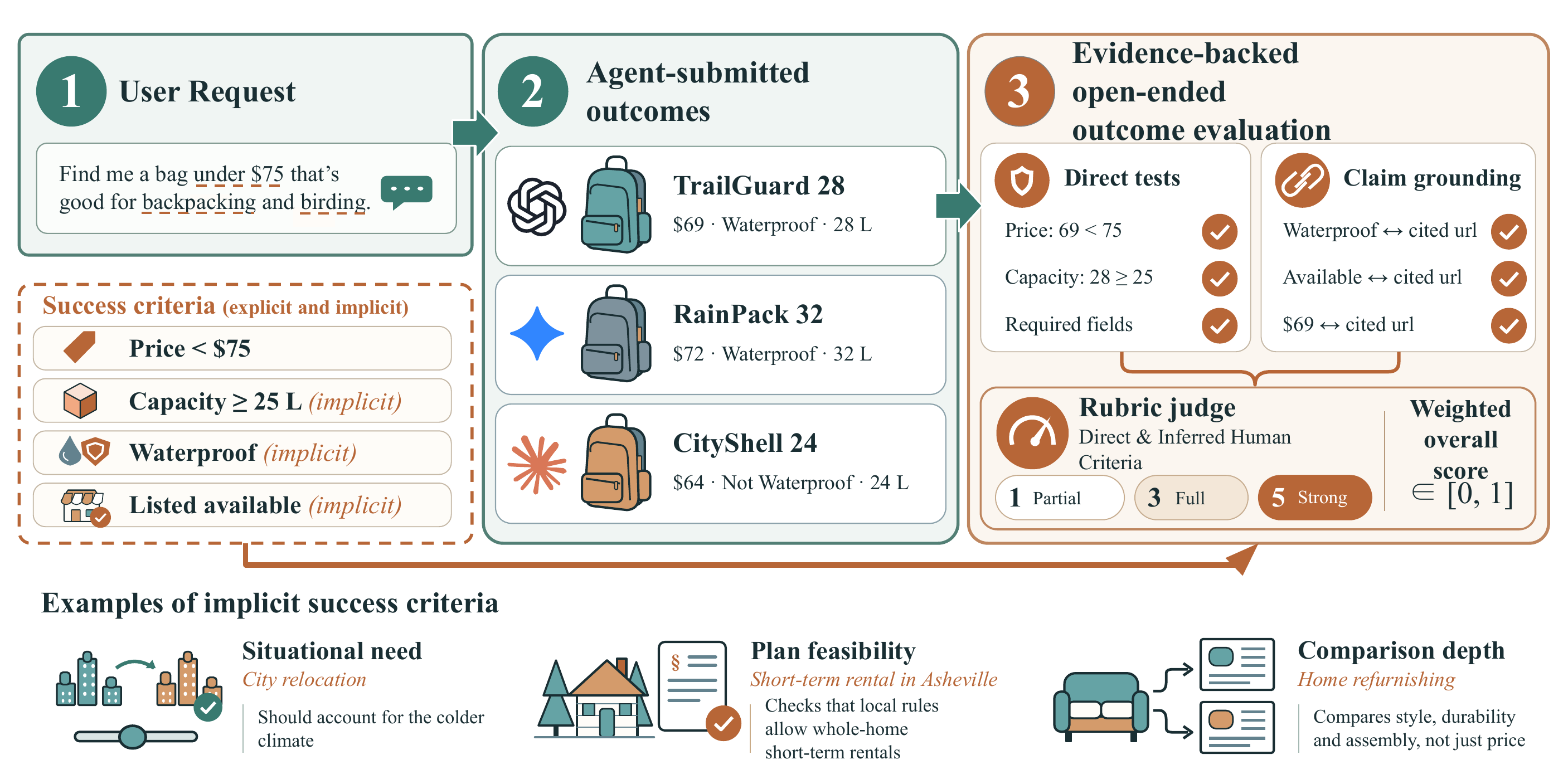}
    \caption{
    Overview of the \benchmarkname framework. 
    A user request is paired with fixed success criteria; agents may submit different outcomes, which are evaluated through deterministic direct tests, source-grounded checks, and rubric-based component scoring.
    }
    \label{fig:overview}
\end{figure}

%% file: sections/1-introduction.tex
\section{Introduction}

AI agents are increasingly being deployed across a wide range of tasks, including everyday digital work such as comparing products, planning trips, gathering information, and preparing user-facing artifacts \citep{yoran2024assistantbenchwebagentssolve, zhou2024webarena, mialon2023gaiabenchmarkgeneralai, he2024webvoyager, yao2024tau, wei2025browsecomp, patwardhan2026gdpval, huang2026therapygym}. 
In these settings, performing relevant actions or retrieving useful information does not necessarily mean that the task has been completed satisfactorily. 
For example, a comparison may omit a requested alternative, a recommendation may rely on unsupported claims, or a travel itinerary may leave out essential logistical details. 
Such outputs may appear complete while leaving important parts of the user's request unresolved. 
These shortcomings raise a fundamental question for agent evaluation: \textbf{\textcolor{BrandPrimary}{``when should an agent's delivered work count as completed?''}}

We study this question through \textbf{\textcolor{BrandPrimary}{satisfactory task completion}}, an outcome-level assessment of whether an agent's delivered work adequately fulfills a delegated request.
Under this standard, an outcome should satisfy the stated requirements, provide a deliverable usable for its intended purpose, and adequately support material claims requiring external evidence.
This notion concerns observable properties of the work product, rather than a user's subjective satisfaction or unexpressed preferences.
Therefore, it provides an operational target for evaluating an agent's work against a task-specific completion standard.

Evaluating satisfactory completion is challenging because everyday requests may admit multiple acceptable outcomes, whose validity can also depend on changing external information. 
Consider the request in Figure \ref{fig:overview}: find a bag under \$75 suitable for full-day backpacking.
Different products may satisfy these requirements, while changes in price or availability can alter which recommendations remain valid.
In this setting, a predetermined reference answer may overlook valid alternatives, while a single overall quality judgment may not reveal specific shortcomings, such as violated constraints, unsupported claims, or missing deliverable components.
Therefore, evaluating satisfactory completion requires assessing the agent's selected outcome against task-specific requirements and supporting evidence, without prescribing the outcome in advance.

Existing agent benchmarks adopt different evaluation designs to make task completion measurable and reproducible.
Mind2Web evaluates actions against recorded demonstrations, while WebArena and OSWorld assess task completion in reproducible environments through executable checks~\citep{deng2023mind2web,zhou2024webarena,xie2024osworld}.
Other benchmarks evaluate realistic user requests against reference answers or use task-specific programs to verify final outputs, often favoring relatively stable information for reliable evaluation~\citep{yoran2024assistantbenchwebagentssolve,mialon2023gaiabenchmarkgeneralai,wei2025browsecomp,cocoabench2026}.
These approaches provide verifiable measures of success, but do not necessarily assess whether agent-selected outcomes remain valid as external information changes.
Meanwhile, Odysseys evaluates live-web workflows through graded rubric checkpoints~\citep{odysseys2026}, while ResearchRubrics and LiveResearchBench use fine-grained criteria to assess open-ended, evidence-backed research outputs~\citep{sharma2025researchrubricsbenchmarkpromptsrubrics,wang2026liveresearchbenchlivebenchmarkusercentric}.
Building on these approaches, we focus on assessing whether diverse, agent-selected outcomes satisfy predefined completion criteria, are supported by current web evidence, and constitute usable deliverables.

To study satisfactory task completion, we introduce \benchmarkname, a benchmark of 77 \textit{human-authored} tasks spanning eight domains of everyday life and work.
Our central design principle is to fix the standard of completion, not the solution.
Each task pairs a natural-language user request with a task-specific success specification established before agent execution.
The specification includes explicit task requirements and anchored component rubrics for assessing whether the resulting work serves its intended purpose, without enumerating acceptable outcomes.
Agents remain free to select their own sources, workflows, and concrete results within the task's stated constraints and deliverable requirements.

To evaluate satisfactory task completion, \benchmarkname uses an evidence-informed, component-level assessment.
We first perform deterministic checks of explicit task requirements and claim-grounding checks that verify submitted claims against agent-cited webpages retrieved at evaluation time~\citep{min2023factscore}.
The resulting findings, together with the submitted outcome and task-specific anchored rubrics, are provided to a component-level judge that assesses each aspect of the deliverable.
Component scores are then aggregated to obtain an overall assessment of task completion.
We evaluate the reliability of grounding and component scoring against independent human annotations~\citep{zheng2023judging,liu2023g}, and further examine evaluator consistency through repeated scoring and controlled grounding failures.

We evaluate 23 agent configurations across all 77 tasks in \benchmarkname.
Claude Code with Opus 5.5 and Fable 5.1 tie for the highest strict pass rate at 54.55\%, while only 32.30\% of all agent--task runs meet the benchmark's completion standard.
%
Failure analysis shows that research accounts for the most deductions even for the high-scoring configurations, often because a claim says more than the page it cites, while low-cost configurations lose more points in every category, most of all through output files that are missing or violate the required format and through user needs that the model does not look into.
These findings reveal a gap between producing plausible task-related information and delivering work that is complete, usable, and adequately supported.

%% file: sections/2-related_work.tex
\section{Related Work}
\label{sec:related_work}

\input{table/related_work_table}

\paragraph{AI agent benchmarks.}
Existing benchmarks increasingly capture realistic tasks that users may delegate to AI agents, but differ in how they define and evaluate successful completion.
AgentBench evaluates agents across diverse interactive environments, while GAIA and AssistantBench evaluate answers to real-world requests against collected references~\citep{liu2025agentbenchevaluatingllmsagents, mialon2023gaiabenchmarkgeneralai, yoran2024assistantbenchwebagentssolve}.
Mind2Web evaluates action predictions against recorded demonstrations, whereas WebArena and OSWorld use reproducible environments and executable checks of task completion~\citep{deng2023mind2web, zhou2024webarena, xie2024osworld}.
More recent benchmarks expand the scope of realistic agent tasks.
MCP-Atlas and Toolathlon evaluate multi-step tasks across tool and application ecosystems using claims-based scoring and executable checks, respectively~\citep{bandi2026mcpatlaslargescalebenchmarktooluse, li2026tooldecathlonbenchmarkinglanguage}.
CocoaBench allows flexible tool composition and evaluates final outputs using task-specific programs, while favoring stable external dependencies for reproducibility~\citep{cocoabench2026}.
Online-Mind2Web and Odysseys extend agent evaluation to live websites, using trajectory-level judgments and graded task rubrics~\citep{onlinemind2web, odysseys2026}.
In contrast, \benchmarkname focuses on evaluating agent-selected, user-facing outcomes whose validity depends on current web information, without prescribing a particular answer or execution workflow.
A more comprehensive comparison is provided in Table~\ref{tab:benchmark_comparison}.

\paragraph{Open-ended outcome evaluation.}
Beyond action-level evaluation, prior work has developed methods for assessing complex outputs that admit multiple valid solutions.
SWE-bench evaluates software patches through executable tests, while PaperBench uses author-validated hierarchical rubrics to assess research replication~\citep{swe-bench2024, starace2025paperbenchevaluatingaisability}.
Agent-as-a-Judge introduces agentic evaluation of AI development tasks against hierarchical user requirements~\citep{agent-as-a-judge2024}.
For web-grounded research, Mind2Web 2 assesses answer correctness and supporting sources through agentic judges, while ResearchRubrics uses expert-defined criteria to evaluate evidence-backed research outputs~\citep{gou2025mind2web2evaluatingagentic, sharma2025researchrubricsbenchmarkpromptsrubrics}.
LiveResearchBench further evaluates dynamic, citation-grounded research reports through complementary content- and report-level assessments~\citep{wang2026liveresearchbenchlivebenchmarkusercentric}.
These works demonstrate the value of evaluating outcomes through explicit requirements, executable verification, and fine-grained rubrics~\citep{arora2025healthbench,du2026deepresearch}.
\benchmarkname builds on these approaches to evaluate a broader range of everyday user-facing outcomes, including product recommendations, travel plans, information collections, and generated artifacts, by jointly assessing constraint satisfaction, current web evidence, and deliverable quality.

%% file: table/related_work_table.tex
\begin{table*}[!t]
\centering
\scriptsize
\caption{
Comparison of representative AI agent benchmarks, grouped by evaluation paradigm.
\textbf{Everyday Tasks} covers users' everyday personal or workplace requests.
\textbf{Real Web} indicates interaction with live real-world websites.
\textbf{Open Outcomes} indicates that evaluation accepts substantively different valid results.
\textbf{Deliverable Form} is the scored output: Answer (short or fixed-field answer), State (environment state), File (edited or generated file), Trajectory (action history and screenshots), Report (long-form text report), and Multi-format (a structured outcome together with user-facing reports, plans, tables, files, or tools).
\textbf{Executable Constraint Checks} uses task-specific deterministic checks of output constraints, environment states, or artifact properties, beyond reference-answer or recorded-action matching alone.
\textbf{Source Verification} assesses whether cited evidence supports submitted claims.
\textbf{Human-Curated Per-Task Rubrics} indicates that human curators author judging rubrics or criteria individually for each task, rather than relying on answer matching, a generic judging prompt, or LLM-generated criteria.
Evaluation columns refer to task-scoring protocols, excluding auxiliary analyses.
\cmark{} = yes, \notcheckmark{} = partial, \xmark{} = no.
}
\label{tab:benchmark_comparison}
\setlength{\tabcolsep}{6pt}
\renewcommand{\arraystretch}{1.12}
\begin{tabular}{@{}lcccc|ccc@{}}
\toprule
\textbf{Benchmark} &
\shortstack{\textbf{Everyday}\\\textbf{Tasks}} &
\shortstack{\textbf{Real}\\\textbf{Web}} &
\shortstack{\textbf{Open}\\\textbf{Outcomes}} &
\shortstack{\textbf{Deliverable}\\\textbf{Form}} &
\shortstack{\textbf{Executable}\\\textbf{Constraint Checks}} &
\shortstack{\textbf{Source}\\\textbf{Verification}} &
\shortstack{\textbf{Human-Curated}\\\textbf{Per-Task Rubrics}} \\
\midrule
\multicolumn{8}{@{}l}{\textit{Reference-answer web assistants}} \\
\quad GAIA \citep{mialon2023gaiabenchmarkgeneralai} & \notcheckmark & \cmark & \xmark & Answer & \xmark & \xmark & \xmark \\
\quad AssistantBench \citep{yoran2024assistantbenchwebagentssolve} & \notcheckmark & \cmark & \xmark & Answer & \xmark & \xmark & \xmark \\
\addlinespace[2pt]
\multicolumn{8}{@{}l}{\textit{Sandboxed execution environments}} \\
\quad WebArena \citep{zhou2024webarena} & \notcheckmark & \xmark & \notcheckmark & Answer/State & \cmark & \xmark & \xmark \\
\quad OSWorld \citep{xie2024osworld} & \cmark & \notcheckmark & \notcheckmark & State/File & \cmark & \xmark & \xmark \\
\addlinespace[2pt]
\multicolumn{8}{@{}l}{\textit{Deep-research reports}} \\
\quad ResearchRubrics \citep{sharma2025researchrubricsbenchmarkpromptsrubrics} & \notcheckmark & \cmark & \cmark & Report & \xmark & \notcheckmark & \cmark \\
\quad LiveResearchBench \citep{wang2026liveresearchbenchlivebenchmarkusercentric} & \notcheckmark & \cmark & \cmark & Report & \xmark & \cmark & \notcheckmark \\
\addlinespace[2pt]
\multicolumn{8}{@{}l}{\textit{Live-web agent tasks}} \\
\quad CocoaBench \citep{cocoabench2026} & \notcheckmark & \notcheckmark & \xmark & Multi-format & \cmark & \xmark & \xmark \\
\quad Odysseys \citep{odysseys2026} & \cmark & \cmark & \notcheckmark & Trajectory & \xmark & \xmark & \notcheckmark \\
\quad Mind2Web 2 \citep{gou2025mind2web2evaluatingagentic} & \cmark & \cmark & \cmark & Answer & \notcheckmark & \cmark & \cmark \\
\midrule
\multicolumn{8}{@{}l}{\textit{Everyday delegated tasks}} \\
\quad \textbf{\benchmarkname} (Ours) & \cmark & \cmark & \cmark & \textbf{Multi-format} & \cmark & \cmark & \cmark \\
\bottomrule
\end{tabular}
\end{table*}

%% file: sections/3-benchmark_construction.tex
\section{\benchmarkname}

\subsection{Benchmark Tasks}

\paragraph{Task scope and composition.} 
\benchmarkname comprises 77 \textit{human-authored} tasks framed as complete user-facing goals across eight application domains: consumer products and shopping; travel and local services; food, culture and entertainment; research; education and careers; business; health and public affairs; software and data work.
We classify tasks by their primary goal: lookup and extraction, comparison and selection, planning, analysis and audit, and deliverables and tools. 
Tasks may combine several activities; the primary goal captures what the user ultimately requests rather than every step needed to finish it.

\paragraph{Task validation.}
For each task, we manually reviewed the instruction and evaluation specification alongside pilot runs. 
We checked that the instruction clearly stated the user’s goal and deliverables, that the success criteria captured the principal requirements, and that the structured outcome schema could express the information needed to assess them. 
We ran task-specific direct tests on both positive and negative cases, and checked whether grounding recovered task-relevant content from web pages captured during execution. 
We also reviewed whether the rubric covered the success criteria, evidence assembled across sources, and qualitative aspects of the deliverable. 
These checks assessed whether each task could be represented and evaluated.

\subsection{Open-Outcome Evaluation}
\label{eval-method}
\paragraph{Success specification.} 
We decompose each user request into success criteria: conditions that an acceptable outcome must satisfy, without requiring a match to a fixed reference answer. 
Each user request has 7.6 success criteria on average. 
We also provide an agent-visible outcome schema specifying how agents submit the information needed to check those conditions. 
For example, a studio-rental task lets agents search public listings and recommend one to three properties. 
Its success criteria include \textit{``every selected effective monthly cost and move-in cash value is exactly recomputed and satisfies its cap''} and 
\textit{``each claimed candidate rent quote and availability date is supported by its property-controlled exact listing page.''}
Different sets of recommended properties can satisfy the same success criteria.

\paragraph{Criterion verification.} 
We evaluate each criterion through one of three routes, according to what must be judged: 
(1) deterministic direct tests handle deterministic properties of agent's submission, such as arithmetic, cross-field consistency, and artifact structure.
(2) source-grounded checks determine whether submitted factual claims are supported by the exact web pages the agent cites; an inaccessible page yields an unverified finding rather than a factual rejection.
And (3) rubric-based component scoring measures qualitative judgment or combined evidence. 
In the rental task above, for example, the cost calculation is checked directly, rent and availability claims are grounded in listing pages, and recommendation quality is assessed through the corresponding rubric component.

\paragraph{Component scoring.} 


Each task is divided into a set of predefined outcome components, each scored on a five-point scale. 
For every component, we provide rubric descriptions for scores of 1, 3, and 5: a score of 1 indicates that the component is missing or substantially unsatisfied; a score of 3 indicates partial fulfillment; and a score of 5 specifies what full and satisfactory fulfillment requires. 
Figure~\ref{fig:overview} shows three representative examples of score-5 anchors.
Scores of 2 and 4 represent performance between adjacent anchors. 
At evaluation time, the component-scoring judge applies these rubrics to the submitted outcome and deliverable, together with the findings from deterministic direct test and source-grounded checks. 
The rubrics capture qualities of agent's submitted outcome where individual direct test or grounded checks cannot fully assess, such as whether evidence from several sources supports a coherent recommendation and whether the final deliverable makes the trade-offs usable for the requester. 
The judge scores each component from 1 to 5; an output that violates the required output format scores 0 on every component.
We combine these component scores using a weighted average, then scale the result to \([0, 1]\) to obtain the task-level reward.

\subsection{Curation Process}
\begin{figure}[h]
\centering
\includegraphics[width=.9\linewidth]{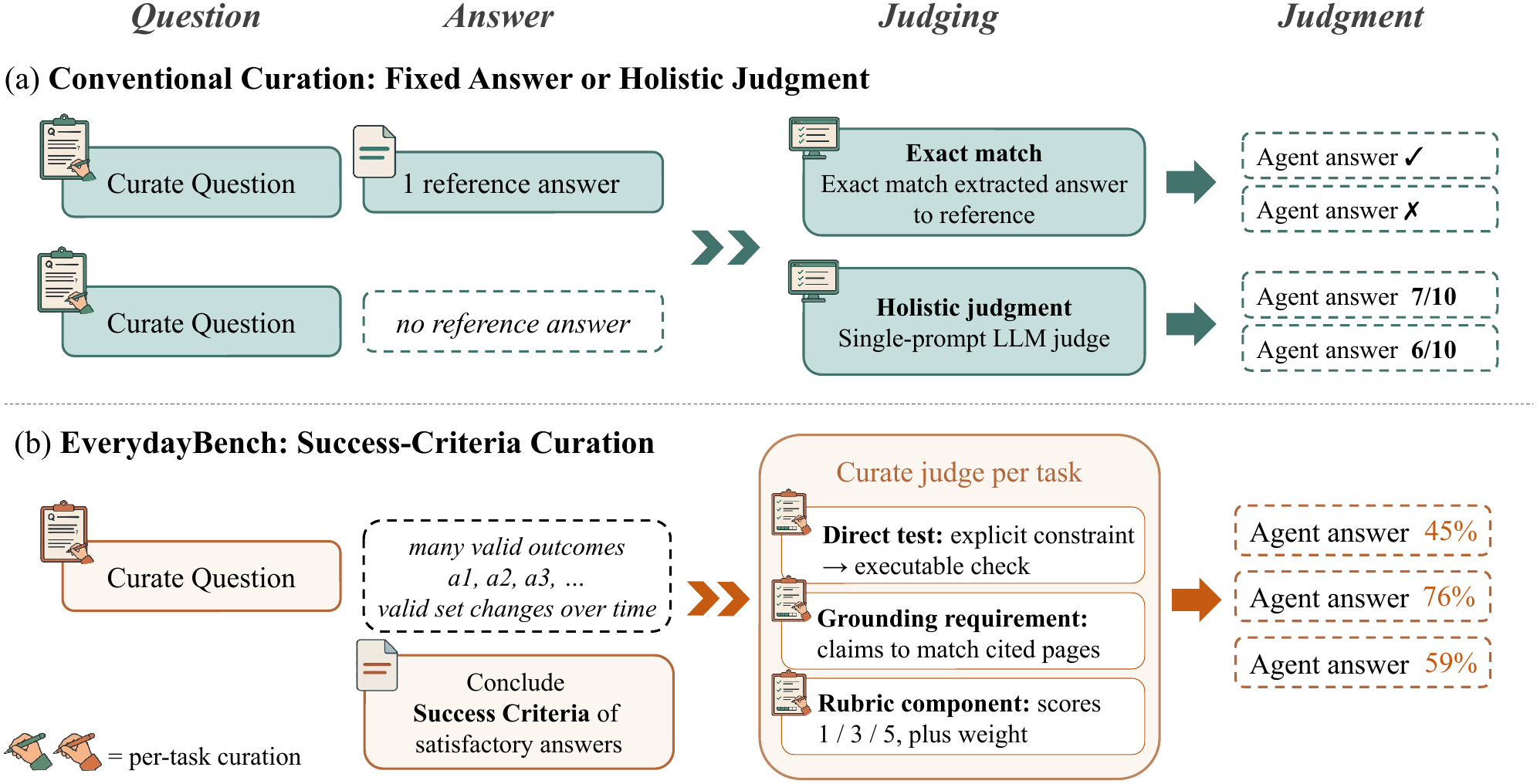}
\caption{Standard fixed-answer curation and our success-criteria curation. Human curators translate an open-ended user request into criterion-level direct tests, grounding requirements, and weighted 1/3/5 rubric anchors, allowing multiple valid outcomes while making specific failures observable.}
\label{fig:curation-process}
\end{figure}

Figure~\ref{fig:curation-process} contrasts conventional fixed-answer or holistic evaluation with our success-criteria curation. Each task is curated by at least three curators, who reach consensus at every step.
First, they derive success criteria from the user's explicit requirements and from the implicit requirements the user would reasonably expect but does not state (e.g., a user relocating from Florida to Minnesota would need guidance on preparing for cold weather).
Second, they assign each criterion to one of three verification routes based on what it requires. Criteria checkable from the submission alone, such as arithmetic, cross-field consistency, or file structure, become \textit{direct tests}. Factual claims that depend on current web content become \textit{source-grounding checks} against the agent's cited pages. Criteria requiring qualitative judgment, such as whether a recommendation weighs the user's trade-offs, are left to the \textit{component rubric}.
Third, they group the criteria into outcome components according to the deliverables the instruction requests, and write the 1/3/5 anchors for each component. Their shared definition of optimal completion becomes the score-5 anchor, the score-3 and score-1 anchors describe progressively incomplete outcomes, and component weights encode relative importance.
Finally, they iterate on positive and negative pilot outcomes, revising criteria and rubrics until valid alternative outcomes are accepted and known failure modes are penalized.


%% file: sections/4-experiments.tex
\section{Evaluation Setup and Validation}
\label{sec:experiment-setting}

\subsection{Evaluation Setup}

We evaluate 23 agent--model configurations on 77 tasks, yielding 1,771 agent runs and corresponding judgments. With official Harbor 0.24.0, each agent is run in an isolated container using a native CLI start and multi-turn resume, with 3,000 seconds allowed per turn. The judging backbone is GPT-5.4 at temperature zero. Agent versions, authentication routes, transports, and resource limits are documented in Appendix~\ref{app:detailed-evaluation-setup}.

Attempts interrupted by timeout or execution failure are excluded and retried until one clean, uninterrupted run completes. Task-specific component scores are weight-averaged by the rubric judge and reported as percentages. A task passes when its overall score is at least 85\% and no rubric component scores below 3; Appendix~\ref{sec:threshold-sensitivity} evaluates sensitivity to the overall-score and rubric-score thresholds. 
Direct-test and grounding findings inform the rubric judge rather than act as separate vetoes, since an isolated finding, such as a cited page that cannot be retrieved, may matter little to the deliverable as a whole. A task can therefore pass despite such a finding if it meets both score thresholds.

\subsection{Evaluation Validation}


\paragraph{Human source-grounding validation.}
Two annotators independently relabeled 90 claim--URL pairs balanced across \textit{matched}, \textit{partial match}, and \textit{not matched}. 
For GPT-5.4, the leaderboard judge, mean human–evaluator agreement was $\kappa= 0.709$, compared with human–human agreement of $\kappa = 0.810$, showing substantial alignment on whether submitted webpages support agent claims.
We use linearly weighted Cohen's $\kappa$ as the three verdicts are ordinal \citep{cohen1968weighted}; full results appear in Appendix~\ref{app:human_validation}.

\paragraph{Human rubric judge validation.}
Two annotators independently scored 72 rubric components from 16 tasks using the same instructions, outcomes, anchors, and verified findings as the evaluator. 
For GPT-5.4, mean human–evaluator agreement was $\kappa = 0.628$, close to human–human agreement of $\kappa= 0.642$. The rubric judge therefore agrees with the annotators about as well as they agree with each other; Appendix~\ref{app:human_validation} gives the full breakdown, including GPT-5.6 Sol as an alternative judge.

\input{figures/temporal_stability_plot}

\paragraph{Scoring stability despite time-sensitive judge.}

\benchmarkname judges each submitted claim against the live version of the page the agent cites, so a score reflects whether an outcome is still valid when it is judged, not only when it was produced. As web content changes, for example when a price, a review count, or an availability window is updated, claims that were correct at submission can lose their support. An outcome built on outdated web content is less useful to the user who acts on it. To examine this sensitivity, we repeatedly evaluated a fixed set of
agent execution traces(the full 77 tasks) after controlled delays (Figure~\ref{fig:temporal-stability}(b)). Because agent outputs are fixed, any change in score reflects changes on the web that our grounding judge successfully captures rather than in agent behavior. The mean score declined from 68.98\% initially to 63.01\% after 20 days and 61.42\% after 30 days.

Despite the temporal sensitivity designed for live-web, relative comparisons of agent performance remained largely stable for benchmarking. We re-ran three Codex agents twice 30 days apart, each with synchronized judging(Figure~\ref{fig:temporal-stability}(a)). In contrast to fixed traces, fresh runs show no systematic decline. These run-to-run changes are also small relative to the gaps between agents, so the ranking by average score remained unchanged. At individual task level, 299 of 342 pairwise orderings (87.4\%) were preserved across evaluations.

\paragraph{Repetition stability.}

Five repeated judge runs of the same fixed 77-task run produced aggregate scores from 85.52\% to 86.26\%, with a 0.24\% standard error for the run mean. Pairwise comparisons revealed no statistically significant differences (\(p=0.611\)), suggesting no detectable run-to-run systematic variation. We also compare our judge with per-task curated rubric with two ablated baselines: \textbf{rubric-only baseline} where our per-task component rubric is used directly as one-pass LLM-as-a-judge rubric; and \textbf{one-prompt baseline} which is one standardized LLM-as-a-judge pass across tasks with full prompt specified in Appendix~\ref{app:judge-ablation-prompts}. Our full evaluator has a significantly more stable mean within-trace SD of 1.41\%, compared with 2.64\% for rubric-only scoring and 4.51\% for one-prompt scoring (Table~\ref{tab:judge-ablation}).

\begin{table}[h]
  \centering
  \small
  \setlength{\tabcolsep}{12pt}
  \caption{Stability of our judge compared with two ablated judges on 50 frozen traces,
  with five judge reruns per trace. Lower standard deviation indicates
  better run-to-run stability.}
  \label{tab:judge-ablation}
  \begin{tabular}{lcc}
    \toprule
    Evaluator variant & Mean reward & Mean SD $\downarrow$ \\
    \midrule
    One-prompt baseline & 0.6096 & 0.0451 \\
    Rubric-only baseline & 0.6671 & 0.0264 \\
    \textbf{Full evaluator (ours)} & 0.5010 & \textbf{0.0141} \\
    \bottomrule
  \end{tabular}
\end{table}

\input{figures/seeded_grounding_barplot}
\paragraph{Grounding under seeded faults.}
Figure~\ref{fig:seeded-grounding-faults} compares our judge to the two baseline judges when we inject 4 types of failure into agent trace. Our full judge successfully detects (when original conclusion of matched changes to partial or unmatched) 85.8\%--97.4\% across each granularity of seeded faults and 90.7\% overall. In comparison, the two baseline judges have lower detection rates of seeded faults when wrong answer is from less relevant contents on the same sites, one piece of value swap (like price), and when entity name is slightly modified. Full per-fault counts appear in Appendix~\ref{app:evaluator-reliability} (Table~\ref{tab:app-seeded-grounding}).

%% file: figures/temporal_stability_plot.tex
\begin{wrapfigure}{r}{0.48\textwidth}
\vspace{-8pt}
\centering
\begin{tikzpicture}[x=0.075cm,y=0.46cm,font=\scriptsize\sffamily]
  \node[anchor=west,font=\scriptsize\sffamily\bfseries,text=EBOrange900]
    at (34.0,6.20) {(a) Fresh execution, synchronized judge};
  \fill[EBOrange300] (48.0,5.45) circle (1.5pt);
  \node[anchor=west,text=black!65] at (49.2,5.45) {initial};
  \fill[EBOrange900] (66.0,5.45) circle (1.5pt);
  \node[anchor=west,text=black!65] at (67.2,5.45) {day 30};

  \foreach \y/\name/\a/\b in {
    4.35/GPT-5.6 Sol/80.17/77.70,
    3.45/GPT-5.6 Terra/77.06/75.99,
    2.55/GPT-5.6 Luna/44.95/49.59} {
      \node[anchor=east,text=black!75] at (37.0,\y) {\name};
      \draw[black!28,line width=0.7pt] (\a,\y)--(\b,\y);
      \fill[EBOrange300] (\a,\y) circle (2.0pt);
      \fill[EBOrange900] (\b,\y) circle (2.0pt);
  }
  \draw[black!35] (38,1.72)--(82,1.72);
  \foreach \x in {40,60,80} {
    \draw[black!35] (\x,1.66)--(\x,1.78);
    \node[anchor=north,text=black!55] at (\x,1.60) {\x};
  }

  \begin{scope}[yshift=-2.85cm]
    \node[anchor=west,font=\scriptsize\sffamily\bfseries,text=EBOrange900]
      at (34.0,6.20) {(b) Fixed outputs, delayed judge};
    \fill[EBOrange300] (40.0,5.45) circle (1.5pt);
    \node[anchor=west,text=black!65] at (41.2,5.45) {initial};
    \fill[EBOrange700] (54.5,5.45) circle (1.5pt);
    \node[anchor=west,text=black!65] at (55.7,5.45) {day 20};
    \fill[EBOrange900] (72.5,5.45) circle (1.5pt);
    \node[anchor=west,text=black!65] at (73.7,5.45) {day 30};

    \foreach \y/\name/\initial/\a/\b in {
      4.55/GPT-5.6 Sol/79.35/73.86/72.43,
      3.75/GPT-5.6 Terra/77.02/71.29/69.83,
      2.95/GPT-5.5/73.58/65.50/62.53,
      2.15/GPT-5.6 Luna/45.97/41.39/40.89} {
        \node[anchor=east,text=black!75] at (37.0,\y) {\name};
        \draw[black!28,line width=0.7pt] (\initial,\y)--(\a,\y)--(\b,\y);
        \fill[EBOrange300] (\initial,\y) circle (2.0pt);
        \fill[EBOrange700] (\a,\y) circle (2.0pt);
        \fill[EBOrange900] (\b,\y) circle (2.0pt);
    }
    \draw[black!35] (38,1.42)--(82,1.42);
    \foreach \x in {40,60,80} {
      \draw[black!35] (\x,1.36)--(\x,1.48);
      \node[anchor=north,text=black!55] at (\x,1.30) {\x};
    }
    \node[anchor=north,text=black!55] at (60,0.58) {mean score (\%)};
  \end{scope}
\end{tikzpicture}
\caption{Panel (a) compares multiple agent runs with near-synchronized judging at different points in time to validate benchmark stability; panel (b) holds agent run fixed to show time-sensitivity to web contents under near-synchronized, 20-day-delayed, and 30-day-delayed judging.}
\label{fig:temporal-stability}
\vspace{-8pt}
\end{wrapfigure}
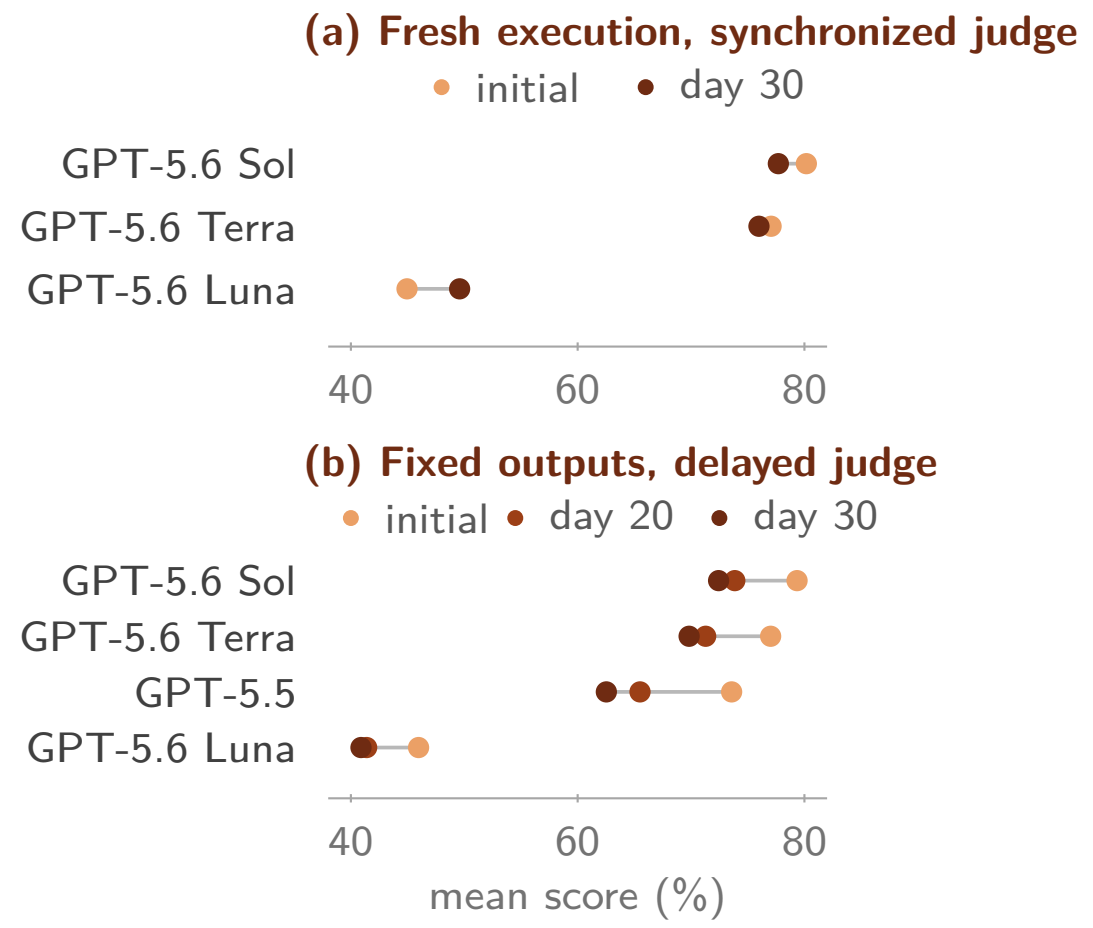

%% file: figures/seeded_grounding_barplot.tex
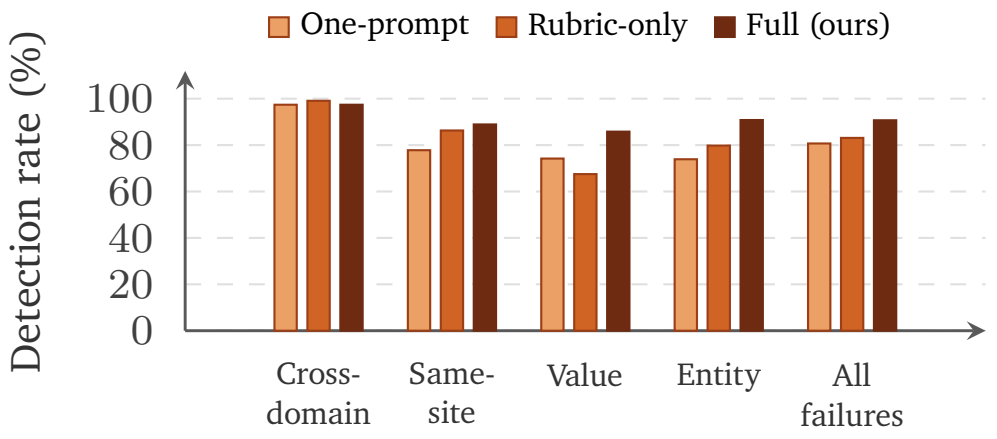
\begin{wrapfigure}{r}{0.4\textwidth}
  \vspace{-10pt}
  \centering
  \centering
  \begin{tikzpicture}
    \begin{axis}[
      ybar,
      width=\linewidth,
      height=0.48\linewidth,
      ymin=0,
      ymax=112,
      ytick={0,20,40,60,80,100},
      ylabel={Detection rate (\%)},
      symbolic x coords={pad-left,Cross-domain,Same-site,Value,Entity,All failures,pad-right},
      xmin=pad-left,
      xmax=pad-right,
      xtick={Cross-domain,Same-site,Value,Entity,All failures},
      xticklabels={
        \shortstack{Cross-\\domain},
        \shortstack{Same-\\site},
        Value,
        Entity,
        \shortstack{All\\failures}
      },
      bar width=4.2pt,
      axis lines=left,
      axis line style={draw=black!65,line width=0.55pt},
      tick style={draw=none},
      ymajorgrids,
      grid style={draw=black!12,dashed},
      tick label style={font=\scriptsize,text=black!75},
      xticklabel style={font=\tiny,align=center,text=black!80},
      label style={font=\scriptsize,text=black!80},
      legend columns=3,
      legend style={
        at={(0.5,1.02)},anchor=south,
        draw=none,fill=none,font=\tiny,
        /tikz/every even column/.append style={column sep=3pt}
      },
      legend image code/.code={
        \draw[#1] (0cm,-0.065cm) rectangle (0.13cm,0.065cm);
      },
      legend cell align=left,
      clip=false,
    ]
      \addplot[
        fill=EBOrange300,
        draw=EBOrange700,
        line width=0.4pt
      ] coordinates {
        (Cross-domain,97.4)
        (Same-site,77.8)
        (Value,74.2)
        (Entity,73.9)
        (All failures,80.7)
      };
      \addlegendentry{One-prompt}

      \addplot[
        fill=EBOrange500,
        draw=EBOrange700,
        line width=0.4pt
      ] coordinates {
        (Cross-domain,99.1)
        (Same-site,86.3)
        (Value,67.5)
        (Entity,79.8)
        (All failures,83.1)
      };
      \addlegendentry{Rubric-only}

      \addplot[
        fill=EBOrange900,
        draw=EBOrange900,
        line width=0.4pt
      ] coordinates {
        (Cross-domain,97.4)
        (Same-site,88.9)
        (Value,85.8)
        (Entity,90.8)
        (All failures,90.7)
      };
      \addlegendentry{Full (ours)}
    \end{axis}
  \end{tikzpicture}
  \caption{Seeded-fault detection by judges.}
  \label{fig:seeded-grounding-faults}
  \vspace{2pt}
\end{wrapfigure}

%% file: sections/5-main_results.tex
\section{Main Results}
\label{sec:main_results}

\subsection{Agent Leaderboard}

Table~\ref{tab:agent-performance} reports strict pass rates for 23 agent--model configurations on \benchmarkname. A task passes only when its overall score is at least 85\% and every rubric component is at least 3. Claude Code with Claude Opus 5.5 and Claude Fable 5.1 tie for first at 54.55\%. Codex with GPT-6.1 Sol and GPT-6 Astra form the next group at 49.35\%, followed by Claude Sonnet 5.5 at 48.05\%. Across all 1,771 agent--task runs, only 32.30\% pass, showing that consistently complete outcomes remain uncommon even when average-quality scores are high.

\begin{table}[t]
\centering
\scriptsize
\setlength{\tabcolsep}{2pt}
\arrayrulecolor{EBOrange900}
\rowcolors{2}{EBOrange050}{white}
\resizebox{\linewidth}{!}{%
\begin{tabular}{@{}lll|cccccc|c@{}}
\hiderowcolors
\toprule
\rowcolor{EBOrange100}
\textbf{Agent} &
\textbf{Model} &
\textbf{Effort} &
\makecell{\textbf{Overall pass}\\\textbf{(n=77)}} &
\makecell{\textbf{Lookup}\\\textbf{(n=16)}} &
\makecell{\textbf{Compare}\\\textbf{(n=25)}} &
\makecell{\textbf{Plan}\\\textbf{(n=12)}} &
\makecell{\textbf{Analysis}\\\textbf{(n=11)}} &
\makecell{\textbf{Build}\\\textbf{(n=13)}} &
\makecell{\textbf{Runtime}\\\textbf{mean $\pm$ SD}\\\textbf{(m:ss)}} \\
\showrowcolors
\midrule
\rowcolor{EBOrange100}
\multicolumn{10}{@{}l}{\textbf{Primary leaderboard}} \\
\rowcolor{EBOrange100}
Claude Code & Claude Opus 5.5 & High & \textbf{54.55} & 50.00 & \textbf{48.00} & 58.33 & \textbf{54.55} & 69.23 & 7:07 $\pm$ 5:21 \\
Claude Code & Claude Fable 5.1 & High & \textbf{54.55} & 43.75 & \textbf{48.00} & \textbf{66.67} & \textbf{54.55} & 69.23 & 8:21 $\pm$ 6:16 \\
Codex     & GPT-6.1 Sol   & High   & 49.35 & 50.00 & 44.00 & 58.33 & \textbf{54.55} & 46.15 & 6:27 $\pm$ 4:18 \\
Codex     & GPT-6 Astra   & High   & 49.35 & 50.00 & 36.00 & 58.33 & 36.36 & \textbf{76.92} & 5:31 $\pm$ 4:06 \\
Claude Code & Claude Sonnet 5.5 & High & 48.05 & 50.00 & 32.00 & 50.00 & 45.45 & \textbf{76.92} & 4:04 $\pm$ 3:23 \\
Codex     & GPT-5.6 Sol   & High   & 42.86 & \textbf{56.25} & 28.00 & 50.00 & 18.18 & 69.23 & 6:34 $\pm$ 12:32 \\
Codex     & GPT-6 Sol     & High   & 38.96 & 50.00 & 28.00 & 58.33 & 18.18 & 46.15 & 4:38 $\pm$ 3:02 \\
Codex     & GPT-5.6 Luna  & High   & 32.47 & 50.00 & 16.00 & 50.00 & 18.18 & 38.46 & 3:04 $\pm$ 2:07 \\
Qwen Code & GPT-6 Sol     & Medium     & 32.47 & 31.25 & 24.00 & 41.67 & 27.27 & 46.15 & 6:30 $\pm$ 18:02 \\
Codex     & GPT-5.6 Terra & High   & 29.87 & 37.50 & 16.00 & 50.00 & 18.18 & 38.46 & 5:25 $\pm$ 11:30 \\
Codex     & GPT-6 Luna    & High   & 29.87 & 25.00 & 20.00 & 33.33 & 45.45 & 38.46 & 2:48 $\pm$ 2:17 \\
Claude Code & Claude Haiku 5.5 & High & 24.68 & 31.25 & 8.00 & 33.33 & 9.09 & 53.85 & 3:52 $\pm$ 2:18 \\
Codex     & GPT-5.5       & High   & 22.08 & 25.00 & 12.00 & 25.00 & 27.27 & 30.77 & 6:06 $\pm$ 2:40 \\
Goose     & GPT-5.6 Luna  & Medium     & 22.08 & 37.50 & 8.00 & 50.00 & 9.09 & 15.38 & 2:11 $\pm$ 1:08 \\
Pi        & GPT-5.6 Luna  & Medium     & 20.78 & 37.50 & 8.00 & 41.67 & 18.18 & 7.69 & 1:53 $\pm$ 1:53 \\
Codex     & GPT-5.5       & Medium & 18.18 & 25.00 & 12.00 & 25.00 & 9.09 & 23.08 & 4:36 $\pm$ 2:35 \\
Qwen Code & GPT-5.6 Luna  & Medium     & 12.99 & 25.00 & 4.00 & 25.00 & 0.00 & 15.38 & 4:33 $\pm$ 18:50 \\
Codex     & GPT-5.5       & Low    & 10.39 & 12.50 & 8.00 & 25.00 & 9.09 & 0.00 & 2:57 $\pm$ 1:36 \\
Qwen Code & Qwen3.7-Plus  & --     & 9.09 & 0.00 & 12.00 & 25.00 & 0.00 & 7.69 & 13:23 $\pm$ 28:59 \\
\midrule
\rowcolor{EBOrange100}
\multicolumn{10}{@{}l}{\textbf{Controlled Codex/GPT-6 Sol reasoning-effort sweep}} \\
Codex     & GPT-6 Sol     & Low    & 29.87 & 31.25 & 16.00 & 50.00 & 18.18 & 46.15 & 2:03 $\pm$ 1:07 \\
Codex     & GPT-6 Sol     & Medium & 28.57 & 25.00 & 24.00 & 33.33 & 27.27 & 38.46 & 3:34 $\pm$ 2:04 \\
Codex     & GPT-6 Sol     & High   & 38.96 & 50.00 & 28.00 & 58.33 & 18.18 & 46.15 & 4:38 $\pm$ 3:02 \\
Codex     & GPT-6 Sol     & XHigh  & 36.36 & 43.75 & 20.00 & 50.00 & 27.27 & 53.85 & 6:15 $\pm$ 3:25 \\
Codex     & GPT-6 Sol     & Max    & 45.45 & 43.75 & 32.00 & 58.33 & \textbf{54.55} & 53.85 & 8:52 $\pm$ 5:32 \\
\bottomrule
\end{tabular}
}
\rowcolors{2}{white}{white}
\arrayrulecolor{black}
\caption{Strict pass rates (\%) across 23 agent--model configurations and five task categories. A task passes when its overall score is at least 85\% and every rubric component is at least 3. Bold marks the best pass rate in each column. 
Attempts interrupted by an execution error or timeout were rerun until one clean run completed; missing or invalid outputs from completed runs score zero.
}
\label{tab:agent-performance}
\end{table}

The task-type pass rates show that no system dominates every kind of everyday work. GPT-5.6 Sol leads lookup at 56.25\%; Opus 5.5 and Fable 5.1 tie on comparison at 48.00\%; Fable leads planning at 66.67\%; and Opus, Fable, GPT-6.1 Sol, and GPT-6 Sol at max effort tie on analysis at 54.55\%. GPT-6 Astra and Sonnet lead customized deliverables at 76.92\%, despite lower overall pass rates than the two leading Claude configurations. The aggregate ranking therefore compresses distinct strengths that matter when an application emphasizes a particular type of work.

Reasoning effort helps most on analysis, but the effect is not monotonic at every adjacent setting. In the controlled GPT-6 Sol sweep, overall pass rate rises from 29.87\% at low effort to 45.45\% at max effort. Analysis pass rate rises from 18.18\% to 54.55\%, while comparison rises from 16.00\% to 32.00\% and lookup from 31.25\% to 43.75\%. Medium and xhigh occasionally trail their neighboring settings, so additional reasoning improves the upper envelope of reliability rather than guaranteeing that every increase in effort produces more completed tasks.

\subsection{Agentic Harness Comparison}
\label{sec:harness-failure-modes}

The agentic harness matters substantially even with the backbone fixed~\citep{jiang2026demystifying}. With GPT-5.6 Luna, Codex passes 32.47\% of tasks, compared with 22.08\% for Goose, 20.78\% for Pi, and 12.99\% for Qwen Code. Codex also leads on lookup at 50.00\% and customized deliverables at 38.46\%, whereas the other three harnesses reach at most 37.50\% and 15.38\%, respectively. With GPT-6 Sol, Codex at high effort passes 38.96\% of tasks versus 32.47\% for Qwen Code. What separates harnesses is therefore not only access to information but the ability to reconcile evidence and package it into a valid, usable deliverable. The failure-mode analysis in Section~\ref{sec:failure-modes} identifies the concrete deductions behind these gaps.

\input{sections/failure_mode_analysis_v25}


\input{sections/section_5_4_v77}

%% file: sections/failure_mode_analysis_v25.tex
\subsection{Failure Mode Analysis}
\label{sec:failure-modes}

We analyze where the agents lose points. We compare two groups of configurations: the \emph{frontier} group, which is four high-scoring configurations, and the \emph{low-cost} group, 
six configurations that each have a median cost per task below \$0.15.
\footnote{Frontier: Claude Opus 5.5, Claude Fable 5.1, GPT-6.1 Sol, and GPT-6 Sol at max reasoning effort. Low-cost: Claude Haiku 5.5, GPT-6 Luna, GPT-5.6 Luna in the Goose, Pi and Qwen Code harnesses, and Qwen3.7-Plus.} Each task has a rubric of several deliverable components (e.g., table, text file). The judge scores every component from 1 to 5 and writes a short reason for each score (Section~\ref{eval-method}). We count each component that scores below 5 as one deduction, and we read the judge's reason for it. An output that violates the required format scores 0, and we count each of its components as a deduction. Across the 77 tasks, 57\% of the components score below 5 in the frontier group and 76\% in the low-cost group. The deductions in the frontier group are mostly small: 72\% of them are a score of 4, against 44\% in the low-cost group.

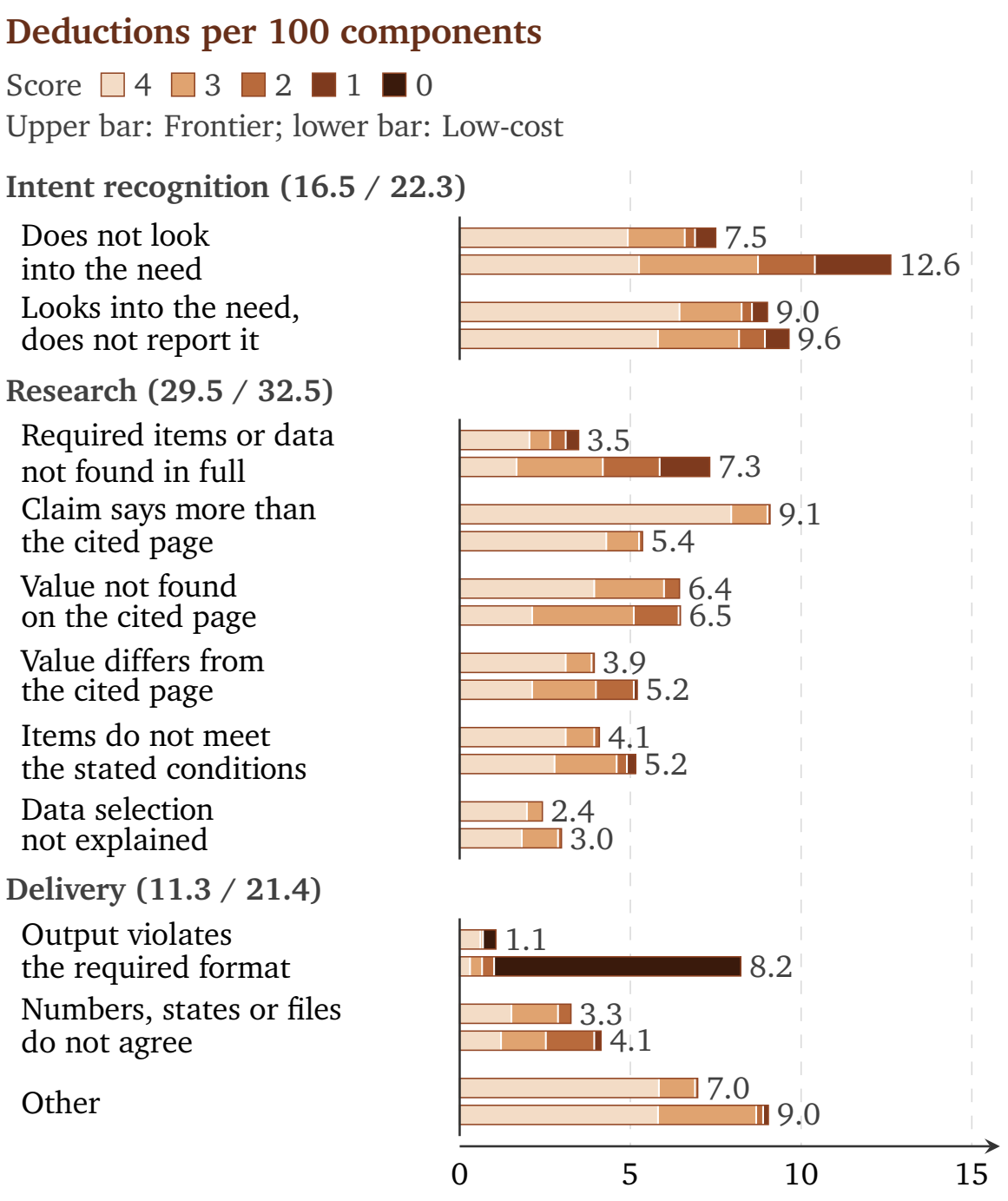
\begin{wrapfigure}{r}{0.46\textwidth}
\vspace{-15pt}
\centering
\resizebox{\linewidth}{!}{\input{sections/fig_deductions_v26_tikz.tex}}
\caption{Deductions per 100 components by failure type for the frontier and low-cost groups over the 77 tasks. A deduction is a component that scores below 5. Numbers in parentheses are the category totals for both groups. 
}
\label{fig:deductions}
\vspace{-10pt}
\end{wrapfigure}

Completing a task takes three things: (1) identifying and working out the user's implicit needs beyond what the request explicitly states (intent recognition), (2) finding the required information (research), and (3) writing what was found into the final files (delivery). We derived the failure types within these categories by hand: we read a random sample of 100 deductions, each with the judge's reason, the agent's output and the agent's trace (the record of the model's messages and tool calls), and grouped the reasons for these deductions into failure types. We then designed a failure-mode judge (Claude Sonnet 5.5) that assigns each deduction one of these failure types. The judge's failure type agreed with the human label on 92 of the 100 hand-labeled deductions, and only then did we use the judge to label all deductions. Appendix~\ref{app:failure-modes} gives the definitions and the prompts. Figure~\ref{fig:deductions} shows the number of deductions per 100 components for each failure type.

\paragraph{Intent recognition.} Intent recognition accounts for 16.5 deductions per 100 components in the frontier group and 22.3 in the low-cost group. 
In these deductions, the output meets the explicit request but misses an implicit need of the user. We further classify these deductions into two failure types by searching the agent's trace and output files: the model does not look into the need, or it looks into the need but does not report it.

\paragraph{The model does not look into the need.} 
This failure type accounts for 7.5 deductions per 100 components in the frontier group and 12.6 in the low-cost group (45\% and 57\% of their intent recognition deductions). The missing point appears neither in the delivered files nor in the model's own messages and tool calls. For example, a user plans to buy a small cabin near Asheville, North Carolina, to operate as a short-term rental, and asks for a quick check of the market first. The implicit need is to check whether renting out a whole home short-term is allowed there, because if it is not, the plan fails however good the market is. GPT-6 Sol (max) notes only that the property must be legally eligible, and does not check the local rules. The model does not check what else must be true for the user's plan to work.

\paragraph{The model looks into the need but does not report it.} This failure type accounts for 9.0 deductions per 100 components in the frontier group and 9.6 in the low-cost group (55\% and 43\% of their intent recognition deductions). The missing point appears in the model's own messages or tool calls but not in the delivered files, or it is stated in a delivered file in a way the user is unlikely to notice. For example, a user asks for wage and growth figures from the US Bureau of Labor Statistics for three AI-related roles, using software developers as the proxy for machine learning engineers. The bureau counts software developers together with quality-assurance analysts and testers, which the user needs to know to read the figures correctly. GPT-6.1 Sol finds this during its research, but the table and summary it delivers do not mention it. The model does not carry what it found into the final files. 


\paragraph{Research.} Research deductions are slightly more frequent in the low-cost group (32.5 per 100 components against 29.5), but the two groups differ in kind. 
Two thirds of the frontier group's research deductions are for values not supported by the cited page, while the low-cost group more often leaves required items missing.


\paragraph{Required items or data are not found in full.} This failure type accounts for 3.5 deductions per 100 components in the frontier group and 7.3 in the low-cost group. For example, asked for a digest of the past week's work on embodied AI drawing on at least three types of source, Qwen3.7-Plus delivers eight items, all of them arXiv papers. For the same task, GPT-6 Sol (max) delivers eight items that cover all four allowed source types: four arXiv papers, two lab blogs from Ai2 and Hugging Face, a technical demo and a conference page. Qwen3.7-Plus stops after a few attempts and reports the shortfall instead of searching further.

\paragraph{The claim says more than the cited page.} This is the most frequent research failure type in the frontier group (9.1 deductions per 100 components against 5.4), and it mostly costs a single point. The cited page supports the core of the claim, but the output adds details the page does not state. For example, asked by a bookstore for sourced tactics to grow a newsletter, Claude Opus 5.5 proposes a referral program with rewards from a shoutout up to store credit; the cited page supports only the shoutout. The model adds plausible details to a sourced claim without marking them as its own.

\paragraph{The value differs from the cited page.} This failure type accounts for 3.9 deductions per 100 components in the frontier group and 5.2 in the low-cost group, and in the low-cost group it also more often costs more than one point. For example, asked for the next meeting of eight machine-learning conferences from their official sites, GPT-5.6 Luna in the Pi harness writes that CVPR 2027 is in Nashville with dates not yet published, while the CVPR 2027 page it cites gives Seattle, June 20 to 25. The model writes a value it already has, here the city of an earlier meeting, without checking it against the cited page.

\paragraph{The value is not found on the cited page.} This failure type occurs at the same rate in both groups (6.4 and 6.5 deductions per 100 components), but in the low-cost group it more often costs more than one point (Figure~\ref{fig:deductions}). The output often cites a general page, such as a home page or a listing. For example, asked for robotics professors each with a current PhD student shown on the lab page, GPT-5.6 Luna in the Goose harness names a student and cites the lab's home page, which shows neither the professor nor the student. The model cites a page on the right topic without checking that the value is on it.

\paragraph{The delivered items do not meet the stated conditions.} This failure type accounts for 4.1 deductions per 100 components in the frontier group and 5.2 in the low-cost group. For example, asked to compare offers for a sneaker from the brand's own store and major retailers, Claude Haiku 5.5 returns five offers, all from the same retailer. The model reaches the required count without checking that the items allow the comparison. 
The remaining research deductions (2.4 and 3.0 per 100 components) are for not explaining how the data were selected, such as which records in a government database were counted for each employer.

\paragraph{Delivery.} Delivery deductions are almost twice as frequent in the low-cost group (21.4 per 100 components against 11.3).


\paragraph{The output violates the required format.} This failure type accounts for 1.1 deductions per 100 components in the frontier group and 8.2 in the low-cost group, where 7.1\% of the runs deliver no valid output file, against 0.3\% in the frontier group: either the file violates the required format, or the agent stops without writing it. For example, on a venture-portfolio task the file that Qwen3.7-Plus delivers lacks a field that the format requires. The model does not validate the file it writes against the required format.

\paragraph{Numbers, states or files do not agree.} The delivered files do not agree with each other, or the times, numbers or states within one file do not fit together. For example, on a task that asks for a workshop built on a specific exhibition at Boston University, GPT-5.6 Luna in the Qwen Code harness names one person from the exhibition page as the speaker in its JSON file and builds its document around a different person. The model writes each part of its output separately without checking that the parts agree. The remaining delivery deductions mostly concern recommendations, plans and trade-off analyses that are not worked out for the user's case, for example workshop activities that could be used with many other photography exhibitions.



%% file: sections/fig_deductions_v26_tikz.tex
\begin{tikzpicture}[x=1cm,y=1cm]
\definecolor{ebScoreFour}{HTML}{F3DCC6}\definecolor{ebScoreThree}{HTML}{E0A370}\definecolor{ebScoreTwo}{HTML}{B86A3C}\definecolor{ebScoreOne}{HTML}{7E3A1E}\definecolor{ebScoreZero}{HTML}{3A1A0C}
\definecolor{ebBrown}{HTML}{672F19}\definecolor{ebRust}{HTML}{914523}\definecolor{ebGrid}{HTML}{D9D9D9}\definecolor{ebAxis}{HTML}{333333}\definecolor{ebText}{HTML}{404040}
\def\ebL{\fontsize{8}{9}\selectfont}\def\ebS{\fontsize{7}{8}\selectfont}
\path[use as bounding box] (0,0) rectangle (8.00,9.410);
\node[anchor=west,inner sep=0pt,font=\ebL\bfseries,text=ebBrown] at (0.000,9.150) {Deductions per 100 components};
\node[anchor=west,inner sep=0pt,font=\ebS,text=ebText] at (0.000,8.750) {Score};
\path[fill=ebScoreFour] (0.750,8.660) rectangle (0.930,8.840);
\draw[ebRust,line width=0.3pt] (0.750,8.660) rectangle (0.930,8.840);
\node[anchor=west,inner sep=0pt,font=\ebS,text=ebText] at (1.000,8.750) {4};
\path[fill=ebScoreThree] (1.300,8.660) rectangle (1.480,8.840);
\draw[ebRust,line width=0.3pt] (1.300,8.660) rectangle (1.480,8.840);
\node[anchor=west,inner sep=0pt,font=\ebS,text=ebText] at (1.550,8.750) {3};
\path[fill=ebScoreTwo] (1.850,8.660) rectangle (2.030,8.840);
\draw[ebRust,line width=0.3pt] (1.850,8.660) rectangle (2.030,8.840);
\node[anchor=west,inner sep=0pt,font=\ebS,text=ebText] at (2.100,8.750) {2};
\path[fill=ebScoreOne] (2.400,8.660) rectangle (2.580,8.840);
\draw[ebRust,line width=0.3pt] (2.400,8.660) rectangle (2.580,8.840);
\node[anchor=west,inner sep=0pt,font=\ebS,text=ebText] at (2.650,8.750) {1};
\path[fill=ebScoreZero] (2.950,8.660) rectangle (3.130,8.840);
\draw[ebRust,line width=0.3pt] (2.950,8.660) rectangle (3.130,8.840);
\node[anchor=west,inner sep=0pt,font=\ebS,text=ebText] at (3.200,8.750) {0 };
\node[anchor=west,inner sep=0pt,font=\ebS,text=ebText] at (0.000,8.410) {Upper bar: Frontier; lower bar: Low-cost};
\draw[ebGrid,dashed,line width=0.3pt] (4.883,8.080) -- (4.883,0.460);
\draw[ebGrid,dashed,line width=0.3pt] (6.217,8.080) -- (6.217,0.460);
\draw[ebGrid,dashed,line width=0.3pt] (7.550,8.080) -- (7.550,0.460);
\node[anchor=west,inner sep=0pt,font=\ebS\bfseries,text=ebText] at (0.000,7.930) {Intent recognition (16.5 / 22.3)};
\node[anchor=west,inner sep=0pt,font=\ebS] at (0.120,7.580) {Does not look};
\node[anchor=west,inner sep=0pt,font=\ebS] at (0.120,7.320) {into the need};
\path[fill=ebScoreFour] (3.550,7.480) rectangle (4.863,7.630);
\path[fill=ebScoreThree] (4.863,7.480) rectangle (5.308,7.630);
\path[fill=ebScoreTwo] (5.308,7.480) rectangle (5.388,7.630);
\path[fill=ebScoreOne] (5.388,7.480) rectangle (5.550,7.630);
\draw[white,line width=0.4pt] (4.863,7.480) -- (4.863,7.630);
\draw[white,line width=0.4pt] (5.308,7.480) -- (5.308,7.630);
\draw[white,line width=0.4pt] (5.388,7.480) -- (5.388,7.630);
\draw[ebRust,line width=0.3pt] (3.550,7.480) rectangle (5.550,7.630);
\node[anchor=west,inner sep=0pt,font=\ebS,text=ebText] at (5.610,7.555) {7.5};
\path[fill=ebScoreFour] (3.550,7.270) rectangle (4.951,7.420);
\path[fill=ebScoreThree] (4.951,7.270) rectangle (5.880,7.420);
\path[fill=ebScoreTwo] (5.880,7.270) rectangle (6.324,7.420);
\path[fill=ebScoreOne] (6.324,7.270) rectangle (6.917,7.420);
\draw[white,line width=0.4pt] (4.951,7.270) -- (4.951,7.420);
\draw[white,line width=0.4pt] (5.880,7.270) -- (5.880,7.420);
\draw[white,line width=0.4pt] (6.324,7.270) -- (6.324,7.420);
\draw[ebRust,line width=0.3pt] (3.550,7.270) rectangle (6.917,7.420);
\node[anchor=west,inner sep=0pt,font=\ebS,text=ebText] at (6.977,7.345) {12.6};
\node[anchor=west,inner sep=0pt,font=\ebS] at (0.120,7.000) {Looks into the need,};
\node[anchor=west,inner sep=0pt,font=\ebS] at (0.120,6.740) {does not report it};
\path[fill=ebScoreFour] (3.550,6.900) rectangle (5.267,7.050);
\path[fill=ebScoreThree] (5.267,6.900) rectangle (5.752,7.050);
\path[fill=ebScoreTwo] (5.752,6.900) rectangle (5.833,7.050);
\path[fill=ebScoreOne] (5.833,6.900) rectangle (5.954,7.050);
\draw[white,line width=0.4pt] (5.267,6.900) -- (5.267,7.050);
\draw[white,line width=0.4pt] (5.752,6.900) -- (5.752,7.050);
\draw[white,line width=0.4pt] (5.833,6.900) -- (5.833,7.050);
\draw[ebRust,line width=0.3pt] (3.550,6.900) rectangle (5.954,7.050);
\node[anchor=west,inner sep=0pt,font=\ebS,text=ebText] at (6.014,6.975) {9.0};
\path[fill=ebScoreFour] (3.550,6.690) rectangle (5.099,6.840);
\path[fill=ebScoreThree] (5.099,6.690) rectangle (5.732,6.840);
\path[fill=ebScoreTwo] (5.732,6.690) rectangle (5.934,6.840);
\path[fill=ebScoreOne] (5.934,6.690) rectangle (6.122,6.840);
\draw[white,line width=0.4pt] (5.099,6.690) -- (5.099,6.840);
\draw[white,line width=0.4pt] (5.732,6.690) -- (5.732,6.840);
\draw[white,line width=0.4pt] (5.934,6.690) -- (5.934,6.840);
\draw[ebRust,line width=0.3pt] (3.550,6.690) rectangle (6.122,6.840);
\node[anchor=west,inner sep=0pt,font=\ebS,text=ebText] at (6.182,6.765) {9.6};
\node[anchor=west,inner sep=0pt,font=\ebS\bfseries,text=ebText] at (0.000,6.350) {Research (29.5 / 32.5)};
\node[anchor=west,inner sep=0pt,font=\ebS] at (0.120,6.000) {Required items or data};
\node[anchor=west,inner sep=0pt,font=\ebS] at (0.120,5.740) {not found in full};
\path[fill=ebScoreFour] (3.550,5.900) rectangle (4.095,6.050);
\path[fill=ebScoreThree] (4.095,5.900) rectangle (4.257,6.050);
\path[fill=ebScoreTwo] (4.257,5.900) rectangle (4.378,6.050);
\path[fill=ebScoreOne] (4.378,5.900) rectangle (4.479,6.050);
\draw[white,line width=0.4pt] (4.095,5.900) -- (4.095,6.050);
\draw[white,line width=0.4pt] (4.257,5.900) -- (4.257,6.050);
\draw[white,line width=0.4pt] (4.378,5.900) -- (4.378,6.050);
\draw[ebRust,line width=0.3pt] (3.550,5.900) rectangle (4.479,6.050);
\node[anchor=west,inner sep=0pt,font=\ebS,text=ebText] at (4.539,5.975) {3.5};
\path[fill=ebScoreFour] (3.550,5.690) rectangle (3.994,5.840);
\path[fill=ebScoreThree] (3.994,5.690) rectangle (4.668,5.840);
\path[fill=ebScoreTwo] (4.668,5.690) rectangle (5.112,5.840);
\path[fill=ebScoreOne] (5.112,5.690) rectangle (5.503,5.840);
\draw[white,line width=0.4pt] (3.994,5.690) -- (3.994,5.840);
\draw[white,line width=0.4pt] (4.668,5.690) -- (4.668,5.840);
\draw[white,line width=0.4pt] (5.112,5.690) -- (5.112,5.840);
\draw[ebRust,line width=0.3pt] (3.550,5.690) rectangle (5.503,5.840);
\node[anchor=west,inner sep=0pt,font=\ebS,text=ebText] at (5.563,5.765) {7.3};
\node[anchor=west,inner sep=0pt,font=\ebS] at (0.120,5.420) {Claim says more than};
\node[anchor=west,inner sep=0pt,font=\ebS] at (0.120,5.160) {the cited page};
\path[fill=ebScoreFour] (3.550,5.320) rectangle (5.671,5.470);
\path[fill=ebScoreThree] (5.671,5.320) rectangle (5.954,5.470);
\path[fill=ebScoreTwo] (5.954,5.320) rectangle (5.974,5.470);
\draw[white,line width=0.4pt] (5.671,5.320) -- (5.671,5.470);
\draw[white,line width=0.4pt] (5.954,5.320) -- (5.954,5.470);
\draw[ebRust,line width=0.3pt] (3.550,5.320) rectangle (5.974,5.470);
\node[anchor=west,inner sep=0pt,font=\ebS,text=ebText] at (6.034,5.395) {9.1};
\path[fill=ebScoreFour] (3.550,5.110) rectangle (4.695,5.260);
\path[fill=ebScoreThree] (4.695,5.110) rectangle (4.951,5.260);
\path[fill=ebScoreTwo] (4.951,5.110) rectangle (4.978,5.260);
\draw[white,line width=0.4pt] (4.695,5.110) -- (4.695,5.260);
\draw[white,line width=0.4pt] (4.951,5.110) -- (4.951,5.260);
\draw[ebRust,line width=0.3pt] (3.550,5.110) rectangle (4.978,5.260);
\node[anchor=west,inner sep=0pt,font=\ebS,text=ebText] at (5.038,5.185) {5.4};
\node[anchor=west,inner sep=0pt,font=\ebS] at (0.120,4.840) {Value not found};
\node[anchor=west,inner sep=0pt,font=\ebS] at (0.120,4.580) {on the cited page};
\path[fill=ebScoreFour] (3.550,4.740) rectangle (4.601,4.890);
\path[fill=ebScoreThree] (4.601,4.740) rectangle (5.146,4.890);
\path[fill=ebScoreTwo] (5.146,4.740) rectangle (5.267,4.890);
\draw[white,line width=0.4pt] (4.601,4.740) -- (4.601,4.890);
\draw[white,line width=0.4pt] (5.146,4.740) -- (5.146,4.890);
\draw[ebRust,line width=0.3pt] (3.550,4.740) rectangle (5.267,4.890);
\node[anchor=west,inner sep=0pt,font=\ebS,text=ebText] at (5.327,4.815) {6.4};
\path[fill=ebScoreFour] (3.550,4.530) rectangle (4.116,4.680);
\path[fill=ebScoreThree] (4.116,4.530) rectangle (4.910,4.680);
\path[fill=ebScoreTwo] (4.910,4.530) rectangle (5.260,4.680);
\path[fill=ebScoreOne] (5.260,4.530) rectangle (5.274,4.680);
\draw[white,line width=0.4pt] (4.116,4.530) -- (4.116,4.680);
\draw[white,line width=0.4pt] (4.910,4.530) -- (4.910,4.680);
\draw[white,line width=0.4pt] (5.260,4.530) -- (5.260,4.680);
\draw[ebRust,line width=0.3pt] (3.550,4.530) rectangle (5.274,4.680);
\node[anchor=west,inner sep=0pt,font=\ebS,text=ebText] at (5.334,4.605) {6.5};
\node[anchor=west,inner sep=0pt,font=\ebS] at (0.120,4.260) {Value differs from};
\node[anchor=west,inner sep=0pt,font=\ebS] at (0.120,4.000) {the cited page};
\path[fill=ebScoreFour] (3.550,4.160) rectangle (4.378,4.310);
\path[fill=ebScoreThree] (4.378,4.160) rectangle (4.580,4.310);
\path[fill=ebScoreTwo] (4.580,4.160) rectangle (4.601,4.310);
\draw[white,line width=0.4pt] (4.378,4.160) -- (4.378,4.310);
\draw[white,line width=0.4pt] (4.580,4.160) -- (4.580,4.310);
\draw[ebRust,line width=0.3pt] (3.550,4.160) rectangle (4.601,4.310);
\node[anchor=west,inner sep=0pt,font=\ebS,text=ebText] at (4.661,4.235) {3.9};
\path[fill=ebScoreFour] (3.550,3.950) rectangle (4.116,4.100);
\path[fill=ebScoreThree] (4.116,3.950) rectangle (4.614,4.100);
\path[fill=ebScoreTwo] (4.614,3.950) rectangle (4.910,4.100);
\path[fill=ebScoreOne] (4.910,3.950) rectangle (4.937,4.100);
\draw[white,line width=0.4pt] (4.116,3.950) -- (4.116,4.100);
\draw[white,line width=0.4pt] (4.614,3.950) -- (4.614,4.100);
\draw[white,line width=0.4pt] (4.910,3.950) -- (4.910,4.100);
\draw[ebRust,line width=0.3pt] (3.550,3.950) rectangle (4.937,4.100);
\node[anchor=west,inner sep=0pt,font=\ebS,text=ebText] at (4.997,4.025) {5.2};
\node[anchor=west,inner sep=0pt,font=\ebS] at (0.120,3.680) {Items do not meet};
\node[anchor=west,inner sep=0pt,font=\ebS] at (0.120,3.420) {the stated conditions};
\path[fill=ebScoreFour] (3.550,3.580) rectangle (4.378,3.730);
\path[fill=ebScoreThree] (4.378,3.580) rectangle (4.601,3.730);
\path[fill=ebScoreTwo] (4.601,3.580) rectangle (4.641,3.730);
\draw[white,line width=0.4pt] (4.378,3.580) -- (4.378,3.730);
\draw[white,line width=0.4pt] (4.601,3.580) -- (4.601,3.730);
\draw[ebRust,line width=0.3pt] (3.550,3.580) rectangle (4.641,3.730);
\node[anchor=west,inner sep=0pt,font=\ebS,text=ebText] at (4.701,3.655) {4.1};
\path[fill=ebScoreFour] (3.550,3.370) rectangle (4.291,3.520);
\path[fill=ebScoreThree] (4.291,3.370) rectangle (4.776,3.520);
\path[fill=ebScoreTwo] (4.776,3.370) rectangle (4.856,3.520);
\path[fill=ebScoreOne] (4.856,3.370) rectangle (4.924,3.520);
\draw[white,line width=0.4pt] (4.291,3.370) -- (4.291,3.520);
\draw[white,line width=0.4pt] (4.776,3.370) -- (4.776,3.520);
\draw[white,line width=0.4pt] (4.856,3.370) -- (4.856,3.520);
\draw[ebRust,line width=0.3pt] (3.550,3.370) rectangle (4.924,3.520);
\node[anchor=west,inner sep=0pt,font=\ebS,text=ebText] at (4.984,3.445) {5.2};
\node[anchor=west,inner sep=0pt,font=\ebS] at (0.120,3.100) {Data selection};
\node[anchor=west,inner sep=0pt,font=\ebS] at (0.120,2.840) {not explained};
\path[fill=ebScoreFour] (3.550,3.000) rectangle (4.075,3.150);
\path[fill=ebScoreThree] (4.075,3.000) rectangle (4.196,3.150);
\draw[white,line width=0.4pt] (4.075,3.000) -- (4.075,3.150);
\draw[ebRust,line width=0.3pt] (3.550,3.000) rectangle (4.196,3.150);
\node[anchor=west,inner sep=0pt,font=\ebS,text=ebText] at (4.256,3.075) {2.4};
\path[fill=ebScoreFour] (3.550,2.790) rectangle (4.035,2.940);
\path[fill=ebScoreThree] (4.035,2.790) rectangle (4.318,2.940);
\path[fill=ebScoreTwo] (4.318,2.790) rectangle (4.345,2.940);
\draw[white,line width=0.4pt] (4.035,2.790) -- (4.035,2.940);
\draw[white,line width=0.4pt] (4.318,2.790) -- (4.318,2.940);
\draw[ebRust,line width=0.3pt] (3.550,2.790) rectangle (4.345,2.940);
\node[anchor=west,inner sep=0pt,font=\ebS,text=ebText] at (4.405,2.865) {3.0};
\node[anchor=west,inner sep=0pt,font=\ebS\bfseries,text=ebText] at (0.000,2.450) {Delivery (11.3 / 21.4)};
\node[anchor=west,inner sep=0pt,font=\ebS] at (0.120,2.100) {Output violates};
\node[anchor=west,inner sep=0pt,font=\ebS] at (0.120,1.840) {the required format};
\path[fill=ebScoreFour] (3.550,2.000) rectangle (3.712,2.150);
\path[fill=ebScoreThree] (3.712,2.000) rectangle (3.732,2.150);
\path[fill=ebScoreZero] (3.732,2.000) rectangle (3.833,2.150);
\draw[white,line width=0.4pt] (3.712,2.000) -- (3.712,2.150);
\draw[white,line width=0.4pt] (3.732,2.000) -- (3.732,2.150);
\draw[ebRust,line width=0.3pt] (3.550,2.000) rectangle (3.833,2.150);
\node[anchor=west,inner sep=0pt,font=\ebS,text=ebText] at (3.893,2.075) {1.1};
\path[fill=ebScoreFour] (3.550,1.790) rectangle (3.631,1.940);
\path[fill=ebScoreThree] (3.631,1.790) rectangle (3.725,1.940);
\path[fill=ebScoreTwo] (3.725,1.790) rectangle (3.819,1.940);
\path[fill=ebScoreZero] (3.819,1.790) rectangle (5.745,1.940);
\draw[white,line width=0.4pt] (3.631,1.790) -- (3.631,1.940);
\draw[white,line width=0.4pt] (3.725,1.790) -- (3.725,1.940);
\draw[white,line width=0.4pt] (3.819,1.790) -- (3.819,1.940);
\draw[ebRust,line width=0.3pt] (3.550,1.790) rectangle (5.745,1.940);
\node[anchor=west,inner sep=0pt,font=\ebS,text=ebText] at (5.805,1.865) {8.2};
\node[anchor=west,inner sep=0pt,font=\ebS] at (0.120,1.520) {Numbers, states or files};
\node[anchor=west,inner sep=0pt,font=\ebS] at (0.120,1.260) {do not agree};
\path[fill=ebScoreFour] (3.550,1.420) rectangle (3.954,1.570);
\path[fill=ebScoreThree] (3.954,1.420) rectangle (4.318,1.570);
\path[fill=ebScoreTwo] (4.318,1.420) rectangle (4.419,1.570);
\draw[white,line width=0.4pt] (3.954,1.420) -- (3.954,1.570);
\draw[white,line width=0.4pt] (4.318,1.420) -- (4.318,1.570);
\draw[ebRust,line width=0.3pt] (3.550,1.420) rectangle (4.419,1.570);
\node[anchor=west,inner sep=0pt,font=\ebS,text=ebText] at (4.479,1.495) {3.3};
\path[fill=ebScoreFour] (3.550,1.210) rectangle (3.873,1.360);
\path[fill=ebScoreThree] (3.873,1.210) rectangle (4.223,1.360);
\path[fill=ebScoreTwo] (4.223,1.210) rectangle (4.601,1.360);
\path[fill=ebScoreOne] (4.601,1.210) rectangle (4.654,1.360);
\draw[white,line width=0.4pt] (3.873,1.210) -- (3.873,1.360);
\draw[white,line width=0.4pt] (4.223,1.210) -- (4.223,1.360);
\draw[white,line width=0.4pt] (4.601,1.210) -- (4.601,1.360);
\draw[ebRust,line width=0.3pt] (3.550,1.210) rectangle (4.654,1.360);
\node[anchor=west,inner sep=0pt,font=\ebS,text=ebText] at (4.714,1.285) {4.1};
\node[anchor=west,inner sep=0pt,font=\ebS] at (0.120,0.810) {Other};
\path[fill=ebScoreFour] (3.550,0.840) rectangle (5.106,0.990);
\path[fill=ebScoreThree] (5.106,0.840) rectangle (5.388,0.990);
\path[fill=ebScoreTwo] (5.388,0.840) rectangle (5.409,0.990);
\draw[white,line width=0.4pt] (5.106,0.840) -- (5.106,0.990);
\draw[white,line width=0.4pt] (5.388,0.840) -- (5.388,0.990);
\draw[ebRust,line width=0.3pt] (3.550,0.840) rectangle (5.409,0.990);
\node[anchor=west,inner sep=0pt,font=\ebS,text=ebText] at (5.469,0.915) {7.0};
\path[fill=ebScoreFour] (3.550,0.630) rectangle (5.099,0.780);
\path[fill=ebScoreThree] (5.099,0.630) rectangle (5.866,0.780);
\path[fill=ebScoreTwo] (5.866,0.630) rectangle (5.920,0.780);
\path[fill=ebScoreOne] (5.920,0.630) rectangle (5.961,0.780);
\draw[white,line width=0.4pt] (5.099,0.630) -- (5.099,0.780);
\draw[white,line width=0.4pt] (5.866,0.630) -- (5.866,0.780);
\draw[white,line width=0.4pt] (5.920,0.630) -- (5.920,0.780);
\draw[ebRust,line width=0.3pt] (3.550,0.630) rectangle (5.961,0.780);
\node[anchor=west,inner sep=0pt,font=\ebS,text=ebText] at (6.021,0.705) {9.0};
\draw[ebAxis,line width=0.5pt] (3.550,7.720) -- (3.550,6.600);
\draw[ebAxis,line width=0.5pt] (3.550,6.140) -- (3.550,2.700);
\draw[ebAxis,line width=0.5pt] (3.550,2.240) -- (3.550,0.540);
\draw[ebAxis,line width=0.5pt,-stealth] (3.550,0.460) -- (7.770,0.460);
\node[anchor=base,inner sep=0pt,font=\ebS] at (3.550,0.160) {0};
\node[anchor=base,inner sep=0pt,font=\ebS] at (4.883,0.160) {5};
\node[anchor=base,inner sep=0pt,font=\ebS] at (6.217,0.160) {10};
\node[anchor=base,inner sep=0pt,font=\ebS] at (7.550,0.160) {15};
\end{tikzpicture}

%% file: sections/section_5_4_v77.tex
\subsection{The Cost-Completion Frontier}
\label{sec:runtime-cost-analysis}
\input{figures/runtime_performance}

Spending more time does not by itself produce more completed tasks. Across the 23 configurations, mean runtime is only moderately associated with pass rate ($\rho=0.45$, $p=0.030$; Figure~\ref{fig:runtime-performance}(a)). Qwen Code with GPT-5.6 Luna runs more than twice as long as Pi with the same model but passes 12.99\% rather than 20.78\% of tasks. Fable 5.1 also runs longer than Opus 5.5 while tying it at 54.55\%. Additional runtime is useful only when the model and harness can turn it into complete outcomes.

Figure~\ref{fig:runtime-performance}(b) converts each trace's uncached input, cache-read, cache-write, and output tokens to a token-only API-equivalent cost using the corresponding public list prices~\cite{openai2026pricing,anthropic2026pricing,alibaba2026pricing}. This normalization does not claim that subscription-authenticated runs incurred an equivalent cash charge. Cost tracks pass rate more closely than runtime does ($\rho=0.58$, $p=0.004$), yet the cost--performance frontier is sparse. Codex with GPT-5.6 Luna passes 32.47\% of tasks at \$0.10 per task; Codex with GPT-6 Sol at high effort costs 8.71$\times$ as much and passes 38.96\%. Opus 5.5 and Fable 5.1 tie at 54.55\%, while Opus uses only 37\% of Fable's API-equivalent token cost.

The controlled GPT-6 Sol sweep makes the compute--completion trade-off explicit. Moving from low to max effort raises pass rate from 29.87\% to 45.45\% while increasing runtime to 4.35$\times$ and cost to 5.06$\times$. The relationship is uneven: moving from high to xhigh raises cost by 22.46\% while pass rate falls from 38.96\% to 36.36\%. Harness choice can matter as much as effort. Qwen Code with GPT-6 Sol costs 1.92$\times$ as much per task as Codex with the same model at high effort, yet passes 32.47\% rather than 38.96\% of tasks. Whether additional tokens translate into completed work is determined by the model and harness, not by budget alone.

%% file: figures/runtime_performance.tex
\definecolor{EBPlotCodex}{HTML}{5B8FD0}
\definecolor{EBPlotClaude}{HTML}{C1654A}
\definecolor{EBPlotQwen}{HTML}{9A80CC}
\definecolor{EBPlotPi}{HTML}{8C95A3}
\definecolor{EBPlotGoose}{HTML}{E0924B}

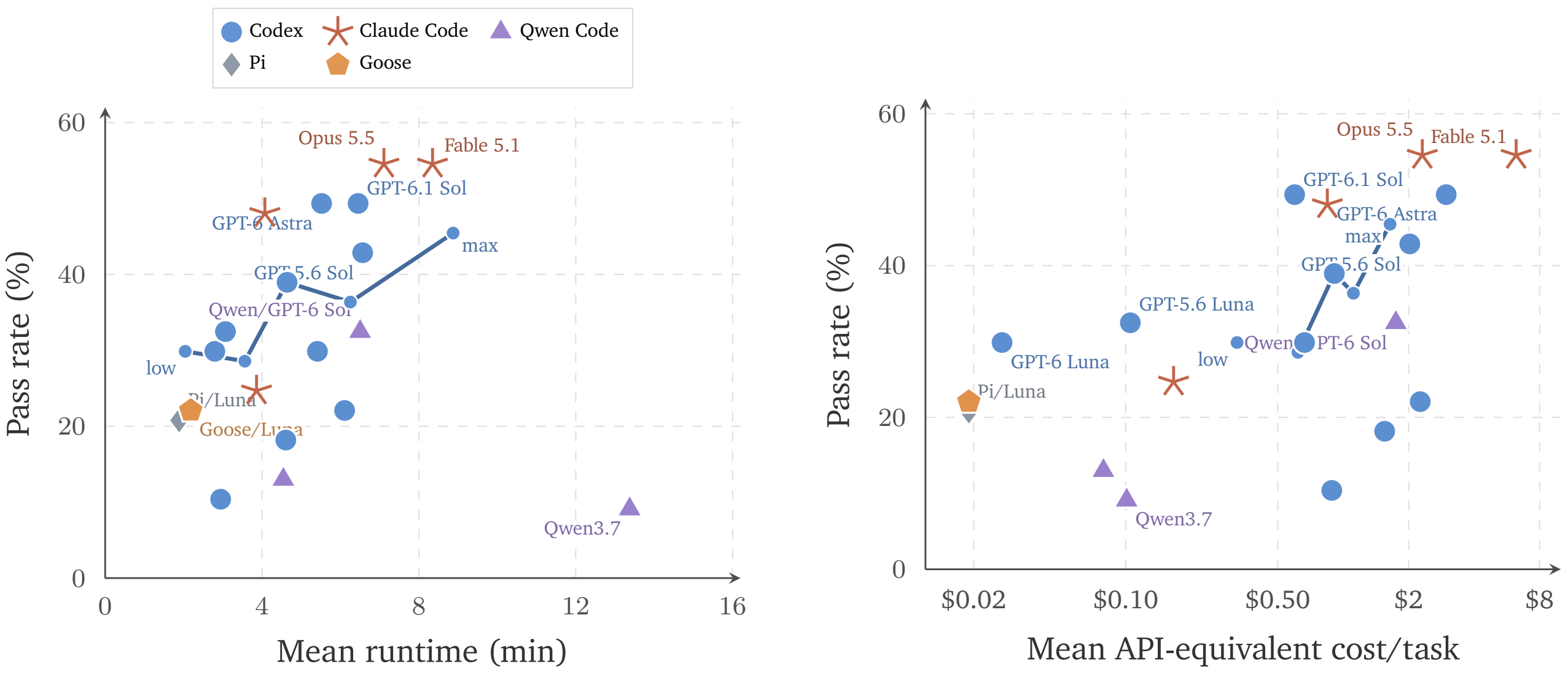
\begin{figure}[h]
\centering
\begin{subfigure}[t]{0.49\linewidth}
\centering
\begin{tikzpicture}
\begin{axis}[
    width=\linewidth,
    height=0.79\linewidth,
    xmin=0, xmax=16.2,
    ymin=0, ymax=62,
    xtick={0,4,8,12,16},
    ytick={0,20,40,60},
    xlabel={Mean runtime (min)},
    ylabel={Pass rate (\%)},
    axis lines=left,
    axis line style={draw=black!70,line width=0.55pt},
    tick style={draw=none},
    grid=both,
    grid style={draw=black!12,dashed},
    tick label style={font=\scriptsize,text=black!70},
    label style={font=\small,text=black!80},
    legend columns=3,
    legend style={
      at={(0.50,1.04)},anchor=south,
      font=\tiny,draw=black!15,fill=white,fill opacity=0.96,
      text opacity=1,inner sep=2.2pt,
      /tikz/every even column/.append style={column sep=4pt}
    },
    legend cell align=left,
    clip=false,
]
\addplot[forget plot,color=EBPlotCodex!75!black,line width=1.25pt,mark=none]
coordinates {(2.042,29.87) (3.563,28.57) (4.640,38.96) (6.254,36.36) (8.874,45.45)};
\addplot[forget plot,only marks,mark=*,mark size=2.2pt,color=EBPlotCodex,
         mark options={draw=white,line width=0.5pt}]
coordinates {(2.042,29.87) (3.563,28.57) (6.254,36.36) (8.874,45.45)};
\addplot[only marks,mark=*,mark size=3.5pt,color=EBPlotCodex,
         mark options={draw=white,line width=0.65pt}]
coordinates {(6.566,42.86) (6.107,22.08) (4.607,18.18) (2.946,10.39) (6.451,49.35) (4.640,38.96) (5.520,49.35) (5.414,29.87) (2.797,29.87) (3.068,32.47)};
\addlegendentry{Codex}
\addplot[only marks,mark=star,mark size=4.9pt,color=EBPlotClaude,
         mark options={draw=EBPlotClaude,line width=0.9pt}]
coordinates {(7.109,54.55) (8.349,54.55) (4.073,48.05) (3.861,24.68)};
\addlegendentry{Claude Code}
\addplot[only marks,mark=triangle*,mark size=4.5pt,color=EBPlotQwen,
         mark options={draw=white,line width=0.65pt}]
coordinates {(6.502,32.47) (13.376,9.09) (4.545,12.99)};
\addlegendentry{Qwen Code}
\addplot[only marks,mark=diamond*,mark size=4.2pt,color=EBPlotPi,
         mark options={draw=white,line width=0.65pt}]
coordinates {(1.887,20.78)};
\addlegendentry{Pi}
\addplot[only marks,mark=pentagon*,mark size=4.2pt,color=EBPlotGoose,
         mark options={draw=white,line width=0.65pt}]
coordinates {(2.182,22.08)};
\addlegendentry{Goose}

\node[font=\tiny,text=EBPlotClaude!80!black,fill=white,inner sep=0.5pt,anchor=south east,xshift=-2pt,yshift=4pt]
  at (axis cs:7.109,54.55) {Opus 5.5};
\node[font=\tiny,text=EBPlotClaude!80!black,fill=white,inner sep=0.5pt,anchor=south west,xshift=3pt,yshift=3pt]
  at (axis cs:8.349,54.55) {Fable 5.1};
\node[font=\tiny,text=EBPlotCodex!80!black,fill=white,inner sep=0.5pt,anchor=north east,xshift=-2pt,yshift=-3pt]
  at (axis cs:6.566,42.86) {GPT-5.6 Sol};
\node[font=\tiny,text=EBPlotCodex!80!black,fill=white,inner sep=0.5pt,anchor=south west,xshift=2pt,yshift=3pt]
  at (axis cs:3.068,32.47) {GPT-5.6 Luna};
\node[font=\tiny,text=EBPlotCodex!80!black,fill=white,inner sep=0.5pt,anchor=north east,xshift=-2pt,yshift=-3pt]
  at (axis cs:5.520,49.35) {GPT-6 Astra};
\node[font=\tiny,text=EBPlotPi!80!black,fill=white,inner sep=0.5pt,anchor=south west,xshift=2pt,yshift=3pt]
  at (axis cs:1.887,20.78) {Pi/Luna};
\node[font=\tiny,text=EBPlotGoose!80!black,fill=white,inner sep=0.5pt,anchor=north west,xshift=2pt,yshift=-3pt]
  at (axis cs:2.182,22.08) {Goose/Luna};
\node[font=\tiny,text=EBPlotQwen!80!black,fill=white,inner sep=0.5pt,anchor=south east,xshift=-2pt,yshift=3pt]
  at (axis cs:6.502,32.47) {Qwen/GPT-6 Sol};
\node[font=\tiny,text=EBPlotQwen!80!black,fill=white,inner sep=0.5pt,anchor=north east,xshift=-2pt,yshift=-3pt]
  at (axis cs:13.376,9.09) {Qwen3.7};
\node[font=\tiny,text=EBPlotCodex!80!black,fill=white,inner sep=0.5pt,anchor=north east,xshift=-2pt,yshift=-2pt]
  at (axis cs:2.042,29.87) {low};
\node[font=\tiny,text=EBPlotCodex!80!black,fill=white,inner sep=0.5pt,anchor=north west,xshift=2pt,yshift=-2pt]
  at (axis cs:8.874,45.45) {max};
\node[font=\tiny,text=EBPlotCodex!80!black,fill=white,inner sep=0.5pt,anchor=south west,xshift=2pt,yshift=2pt]
  at (axis cs:6.451,49.35) {GPT-6.1 Sol};
\end{axis}
\end{tikzpicture}
\caption{Elapsed runtime.}
\label{fig:runtime-panel}
\end{subfigure}
\hfill
\begin{subfigure}[t]{0.49\linewidth}
\centering
\begin{tikzpicture}
\begin{axis}[
    width=\linewidth,
    height=0.79\linewidth,
    xmode=log,
    log basis x=10,
    xmin=0.012, xmax=10,
    ymin=0, ymax=62,
    xtick={0.02,0.1,0.5,2,8},
    xticklabels={\$0.02,\$0.10,\$0.50,\$2,\$8},
    ytick={0,20,40,60},
    xlabel={Mean API-equivalent cost/task},
    ylabel={Pass rate (\%)},
    axis lines=left,
    axis line style={draw=black!70,line width=0.55pt},
    tick style={draw=none},
    grid=both,
    grid style={draw=black!12,dashed},
    tick label style={font=\scriptsize,text=black!70},
    label style={font=\small,text=black!80},
    clip=false,
]
\addplot[forget plot,color=EBPlotCodex!75!black,line width=1.25pt,mark=none]
coordinates {(0.325,29.87) (0.619,28.57) (0.910,38.96) (1.114,36.36) (1.642,45.45)};
\addplot[forget plot,only marks,mark=*,mark size=2.2pt,color=EBPlotCodex,
         mark options={draw=white,line width=0.5pt}]
coordinates {(0.325,29.87) (0.619,28.57) (1.114,36.36) (1.642,45.45)};
\addplot[only marks,mark=*,mark size=3.5pt,color=EBPlotCodex,
         mark options={draw=white,line width=0.65pt}]
coordinates {(2.023,42.86) (2.260,22.08) (1.551,18.18) (0.886,10.39) (0.598,49.35) (0.910,38.96) (2.975,49.35) (0.664,29.87) (0.027,29.87) (0.105,32.47)};
\addplot[only marks,mark=star,mark size=4.9pt,color=EBPlotClaude,
         mark options={draw=EBPlotClaude,line width=0.9pt}]
coordinates {(2.310,54.55) (6.239,54.55) (0.845,48.05) (0.166,24.68)};
\addplot[only marks,mark=triangle*,mark size=4.5pt,color=EBPlotQwen,
         mark options={draw=white,line width=0.65pt}]
coordinates {(1.746,32.47) (0.101,9.09) (0.079,12.99)};
\addplot[only marks,mark=diamond*,mark size=4.2pt,color=EBPlotPi,
         mark options={draw=white,line width=0.65pt}]
coordinates {(0.019,20.78)};
\addplot[only marks,mark=pentagon*,mark size=4.2pt,color=EBPlotGoose,
         mark options={draw=white,line width=0.65pt}]
coordinates {(0.019,22.08)};

\node[font=\tiny,text=EBPlotClaude!80!black,fill=white,inner sep=0.5pt,anchor=south east,xshift=-2pt,yshift=4pt]
  at (axis cs:2.310,54.55) {Opus 5.5};
\node[font=\tiny,text=EBPlotClaude!80!black,fill=white,inner sep=0.5pt,anchor=south east,xshift=-2pt,yshift=3pt]
  at (axis cs:6.239,54.55) {Fable 5.1};
\node[font=\tiny,text=EBPlotCodex!80!black,fill=white,inner sep=0.5pt,anchor=north east,xshift=-2pt,yshift=-3pt]
  at (axis cs:2.023,42.86) {GPT-5.6 Sol};
\node[font=\tiny,text=EBPlotCodex!80!black,fill=white,inner sep=0.5pt,anchor=south west,xshift=2pt,yshift=3pt]
  at (axis cs:0.105,32.47) {GPT-5.6 Luna};
\node[font=\tiny,text=EBPlotCodex!80!black,fill=white,inner sep=0.5pt,anchor=north west,xshift=2pt,yshift=-3pt]
  at (axis cs:0.027,29.87) {GPT-6 Luna};
\node[font=\tiny,text=EBPlotCodex!80!black,fill=white,inner sep=0.5pt,anchor=north east,xshift=-2pt,yshift=-3pt]
  at (axis cs:2.975,49.35) {GPT-6 Astra};
\node[font=\tiny,text=EBPlotPi!80!black,fill=white,inner sep=0.5pt,anchor=south west,xshift=2pt,yshift=3pt]
  at (axis cs:0.019,20.78) {Pi/Luna};
\node[font=\tiny,text=EBPlotQwen!80!black,fill=white,inner sep=0.5pt,anchor=north east,xshift=-2pt,yshift=-3pt]
  at (axis cs:1.746,32.47) {Qwen/GPT-6 Sol};
\node[font=\tiny,text=EBPlotQwen!80!black,fill=white,inner sep=0.5pt,anchor=north west,xshift=2pt,yshift=-3pt]
  at (axis cs:0.101,9.09) {Qwen3.7};
\node[font=\tiny,text=EBPlotCodex!80!black,fill=white,inner sep=0.5pt,anchor=north east,xshift=-2pt,yshift=-2pt]
  at (axis cs:0.325,29.87) {low};
\node[font=\tiny,text=EBPlotCodex!80!black,fill=white,inner sep=0.5pt,anchor=north east,xshift=-2pt,yshift=-2pt]
  at (axis cs:1.642,45.45) {max};
\node[font=\tiny,text=EBPlotCodex!80!black,fill=white,inner sep=0.5pt,anchor=south west,xshift=2pt,yshift=2pt]
  at (axis cs:0.598,49.35) {GPT-6.1 Sol};
\end{axis}
\end{tikzpicture}
\caption{Token-only public-list-price equivalent.}
\label{fig:cost-panel}
\end{subfigure}
\caption{Strict pass rate against runtime and inferred model cost. The connected blue path is the controlled Codex/GPT-6 Sol effort sweep from low to max. Cost combines uncached input, cache reads and writes, and output tokens at public list prices; the logarithmic axis exposes the full price range. Subscription-authenticated runs are normalized to the corresponding API-equivalent rate rather than treated as cash expenditure.}
\label{fig:runtime-performance}
\end{figure}

%% file: sections/6-conclusion.tex
\section{Conclusion}
We introduced \benchmarkname, a benchmark of 77 human-authored tasks for evaluating agent-selected outcomes on the live web.
Each task specifies success criteria before execution without prescribing a single valid answer. 
Evaluation combines direct checks, verification against cited webpages, and component-level scoring of the submitted deliverable.
Across 
23 agent configurations, the strongest agent configuration 
passes 
54.55\%
of tasks.
%
%
Failure mode analysis shows that research accounts for the most deductions in both the high-scoring and the low-cost configurations, and the high-scoring configurations most often lose a point because a claim says more than the page it cites. The low-cost configurations lose more points in every category, most of all through output files that are missing or violate the required format and through user needs that the model does not look into.

%% file: sections/appendix.tex
\definecolor{fmpromptline}{HTML}{4C6F91}
\definecolor{fmpromptback}{HTML}{F4F8FB}
\lstdefinestyle{failuremodeprompt}{
  basicstyle=\ttfamily\fontsize{6.6pt}{8.0pt}\selectfont,
  numbers=left,
  numberstyle=\ttfamily\tiny\color{black!42},
  numbersep=6pt,
  stepnumber=1,
  breaklines=true,
  breakatwhitespace=false,
  columns=fullflexible,
  keepspaces=true,
  showstringspaces=false,
  extendedchars=false,
  frame=none,
  xleftmargin=1.7em,
  xrightmargin=0.4em,
  aboveskip=0pt,
  belowskip=0pt
}
\tcbset{
  failurelisting/prompt/.style={
    enhanced jigsaw,
    breakable,
    listing only,
    listing engine=listings,
    listing options={style=failuremodeprompt},
    width=\linewidth,
    boxrule=0.45pt,
    arc=1mm,
    outer arc=1mm,
    left=0.4mm,
    right=0.4mm,
    top=0.8mm,
    bottom=0.8mm,
    before skip=6pt,
    after skip=7pt,
    fonttitle=\bfseries\small,
    coltitle=black,
    colback=fmpromptback,
    colframe=fmpromptline,
    colbacktitle=fmpromptline!15!white,
    borderline west={1.4pt}{0pt}{fmpromptline}
  }
}

\section{Additional Results}
\label{app:additional-results}

\input{sections/appendix_threshold_sensitivity_v2}

\subsection{Computer Use Versus Retrieval-Only Execution}
\label{sec:computer-use}

We ran Codex with GPT-5.6 Sol on 25  sampled tasks from our benchmark under four web-access settings (Table~\ref{tab:computer-use}): host execution with search only, host execution with computer use, and browser-free containers with either a 900- or 3,000-second budget. All other settings were fixed.

\begin{table}[t]
\caption{Scores (\%) across four web-access settings on the same 25 tasks.}
\label{tab:computer-use}
\begin{center}
\small
\begin{tabular}{lrrr}
\multicolumn{1}{c}{\bf Configuration} & \multicolumn{1}{c}{\bf Browser}
& \multicolumn{1}{c}{\bf Shell network} & \multicolumn{1}{c}{\bf Score} \\
\hline
Host, search only            & no  & blocked & 77.4 \\
Host, computer use           & \bf yes & blocked & \bf 78.4 \\
Container, 3000s budget      & no  & full    & 79.0 \\
Container, 900s budget       & no  & full    & 73.5 \\
\end{tabular}
\end{center}
\end{table}

\begin{figure}[t]
\centering
\includegraphics[width=\textwidth]{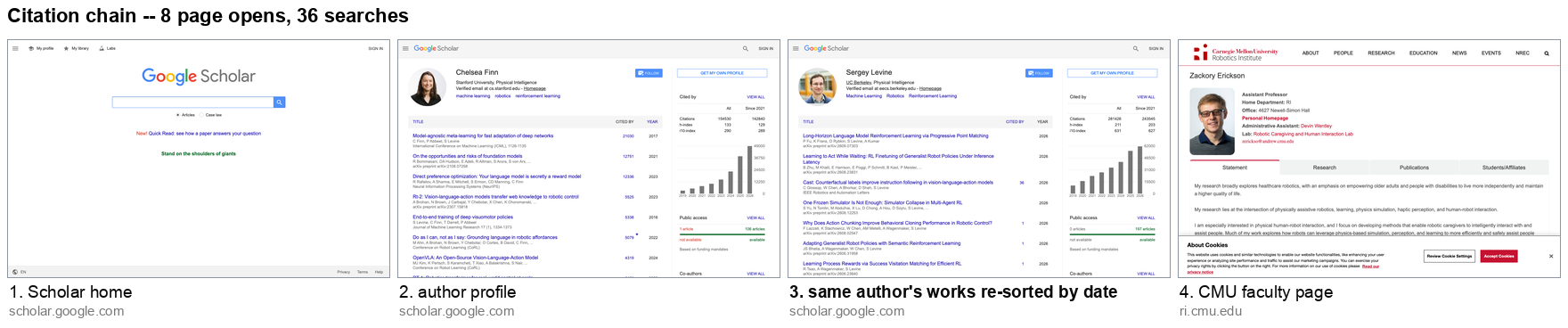}\\[4pt]
\includegraphics[width=\textwidth]{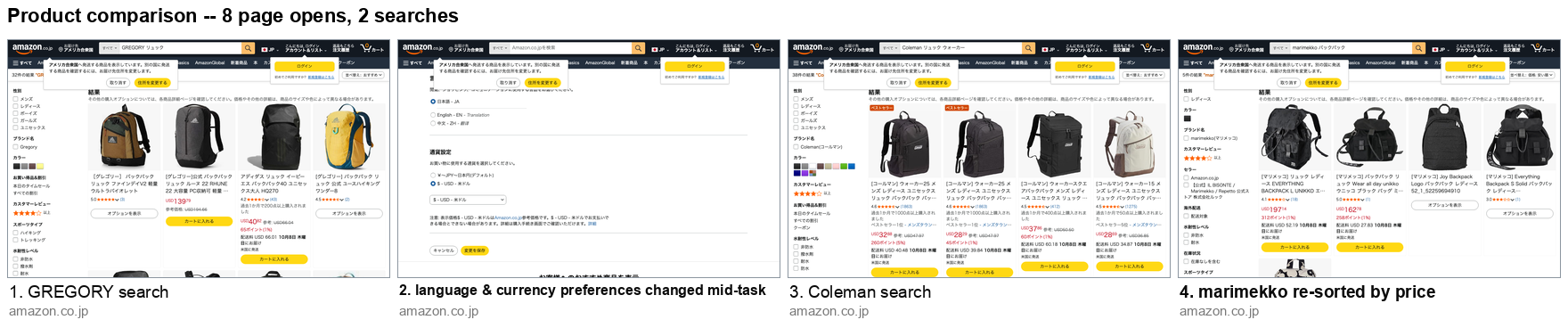}\\[4pt]
\includegraphics[width=\textwidth]{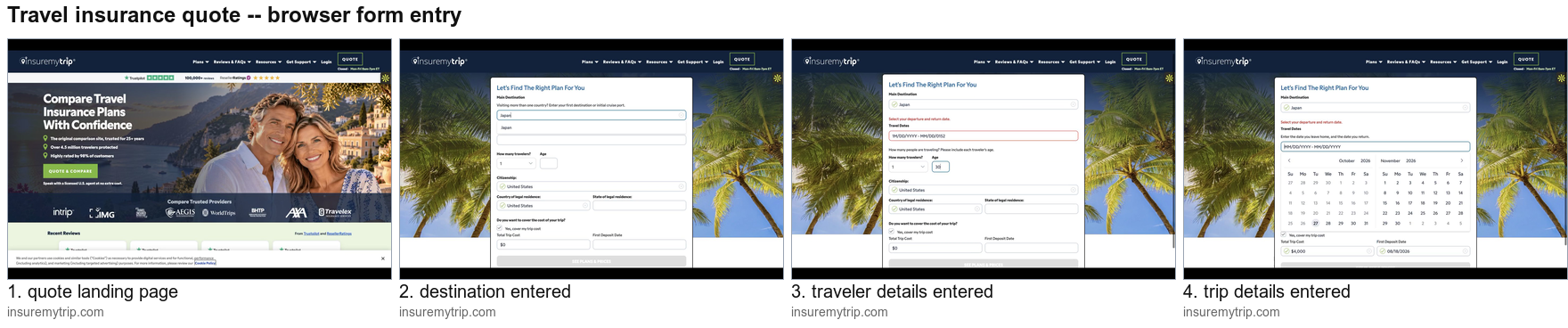}
\caption{Three execution trajectories from the computer-use runs. The final row is
sampled from the InsureMyTrip browser recording and shows the agent progressively
entering quote-search information.}
\label{fig:trajectories}
\end{figure}

Computer use changed how the agent gathered evidence. Representative trajectories (Figure~\ref{fig:trajectories}) include a citation chain that opens author profiles on Google Scholar and re-sorts a publication list by date before following the chain to a university faculty page; a product comparison that changes the store's language and currency settings mid-task and re-sorts a listing by price while working across four brands; and a travel-insurance quote flow that enters destination, traveler, residence, trip-cost, and date information directly into the booking form. A fourth trajectory clears a bot check and triangulates a tariff classification across the Harmonized Tariff Schedule, a CBP ruling, and USTR guidance.

The agent manipulated page state---sorting, filtering, and changing site preferences---rather than retrieving a fixed URL, and the balance between navigation and search varied sharply by task. Yet the changes had little effect on score: the computer-use and search-only runs both yielded a 32.0\% pass rate, with raw rubric scores of 78.4 and 77.4, respectively.

\FloatBarrier
\section{Benchmark and Evaluation Details}
\label{app:benchmark-evaluation-details}

\subsection{Detailed Evaluation Setup}
\label{app:detailed-evaluation-setup}

\paragraph{Agents and evaluator.} We report 23 agent--model configurations and 1,771 agent--task runs, with every configuration covering all 77 tasks. Claude Code uses Claude Opus 5.5, Fable 5.1, Sonnet 5.5, and Haiku 5.5 at high effort. Codex uses GPT-5.5 at low, medium, and high effort; GPT-5.6 Sol, Terra, and Luna at high effort; GPT-6.1 Sol, GPT-6 Astra, Sol, and Luna at high effort; and GPT-6 Sol at low, medium, xhigh, and max effort. Qwen Code uses Qwen3.7-Plus, GPT-5.6 Luna, and GPT-6 Sol, while Pi and Goose use GPT-5.6 Luna. Each agent may choose its own sources, tools, and workflow. The source-grounded checker and rubric-based component scorer use GPT-5.4 at temperature zero,
with one evaluator pass per run.

\paragraph{Configuration fidelity.} Claude Code uses Claude subscription OAuth: the Opus 5.5 run uses version 2.1.311, while the Fable 5.1 traces span the recorded transition from 2.1.305 to 2.1.311. The GPT-5.6 Codex runs use Codex 0.147.0, while the GPT-6 family and effort sweep use Codex 0.163.0; both authenticate with ChatGPT subscription OAuth. Qwen Code 0.26.1, Pi 1.4.0, and Goose v1.56.0 use their native CLI adapters. Qwen3.7-Plus is routed through OpenRouter. Pi/GPT-5.6 Luna, Qwen Code/GPT-5.6 Luna, Goose/GPT-5.6 Luna, and Qwen Code/GPT-6 Sol use the recorded UniAPI OpenAI-compatible route; Pi and Qwen Code/GPT-6 Sol use Responses transport. The latter uses a narrow subclass of Harbor's official Qwen adapter that changes only CLI authentication transport from Chat Completions to Responses. Agent instructions, the native two-step protocol, and the verifier remain unchanged.

\paragraph{Execution environment and protocol.} We orchestrate runs with official Harbor 0.24.0 at suite commit \texttt{f29775a} in isolated Docker containers with internet access. Each task configuration records two CPU cores, 4~GB of memory, 10~GB of storage, and a 3,000-second timeout for each native agent step. Model calls use either the configuration's subscription-authenticated CLI or its recorded API route, while orchestration, artifact collection, and analysis run locally. Harbor first provides the user request and then resumes the same trajectory with the outcome schema used to submit the final structured result (\texttt{resume\_trajectory=true}).

\paragraph{Failure handling and selection.} If a timeout or execution error prevents either native step from completing, we exclude that  attempt and rerun the task. If both steps complete but the required output is missing or cannot be parsed, the evaluator assigns zero. A completed submission that violates the output schema likewise receives zero; these zeros remain because they represent task failures rather than incomplete executions. 

\paragraph{Scoring and aggregation.} We aggregate task-specific component scores using their rubric weights, normalize the result to \([0,1]\), and multiply by 100 for percentage reporting. 
Each configuration contributes one completed run per task. A task passes when it has an overall score
of at least 85\% and every component score of at least 3.



\section{One-Prompt Judge Baseline}
\label{app:judge-ablation-prompts}

The one-prompt baseline tests the stability of a minimally structured holistic evaluator. It receives the exact task instruction, the complete agent result, and the parsed submission summary, then assigns one overall score from 1 to 5 with a short explanation. Unlike our full evaluator, it receives neither task-specific component rubrics nor criterion-level direct-test and grounding findings. The system message and user-message template used for every task are reproduced verbatim below; the angle-bracketed fields are replaced with the corresponding run data at inference time.

\lstdefinestyle{judgeablationprompt}{
  basicstyle=\ttfamily\footnotesize,
  breaklines=true,
  breakindent=0pt,
  columns=fullflexible,
  keepspaces=true,
  showstringspaces=false,
  frame=single,
  framesep=5pt,
  xleftmargin=5pt,
  xrightmargin=5pt
}

\tcbinputlisting{failurelisting/prompt,title={System message},listing file={judge-ablation-assets/one-prompt-system.txt}}

\tcbinputlisting{failurelisting/prompt,title={User-message template},listing file={judge-ablation-assets/one-prompt-user-template.txt}}

\section{Seeded Failure Results}
\label{app:evaluator-reliability}

\begin{table}[h]
\caption{Seeded grounding failures detected by each judge, as count (rate). Dashes mark arms that were not evaluated.}
\label{tab:app-seeded-grounding}
\centering
\footnotesize
\setlength{\tabcolsep}{10pt}
\begin{tabular}{lrccc}
\toprule
\textbf{Fault class} & \textbf{$n$} & \textbf{One-prompt} & \textbf{Rubric-only} & \textbf{Full (ours)} \\
\midrule
Citation swap (cross-domain) & 116 & 113 (97.4\%) & 115 (99.1\%) & 113 (97.4\%) \\
Citation swap (same site)    & 117 &  91 (77.8\%) & 101 (86.3\%) & 104 (88.9\%) \\
Value perturbation           & 120 &  89 (74.2\%) &  81 (67.5\%) & 103 (85.8\%) \\
Entity substitution          & 119 &  88 (73.9\%) &  95 (79.8\%) & 108 (90.8\%) \\
\midrule
All seeded failures          & 472 & 381 (80.7\%) & 392 (83.1\%) & 428 (90.7\%) \\
Control (unmodified)         & 120 & -- & -- & 14 (11.7\%) \\
\bottomrule
\end{tabular}
\end{table}

\section{Human Validation}
\label{app:human_validation}

\begin{figure}[H]
    \centering
    \includegraphics[width=\linewidth]{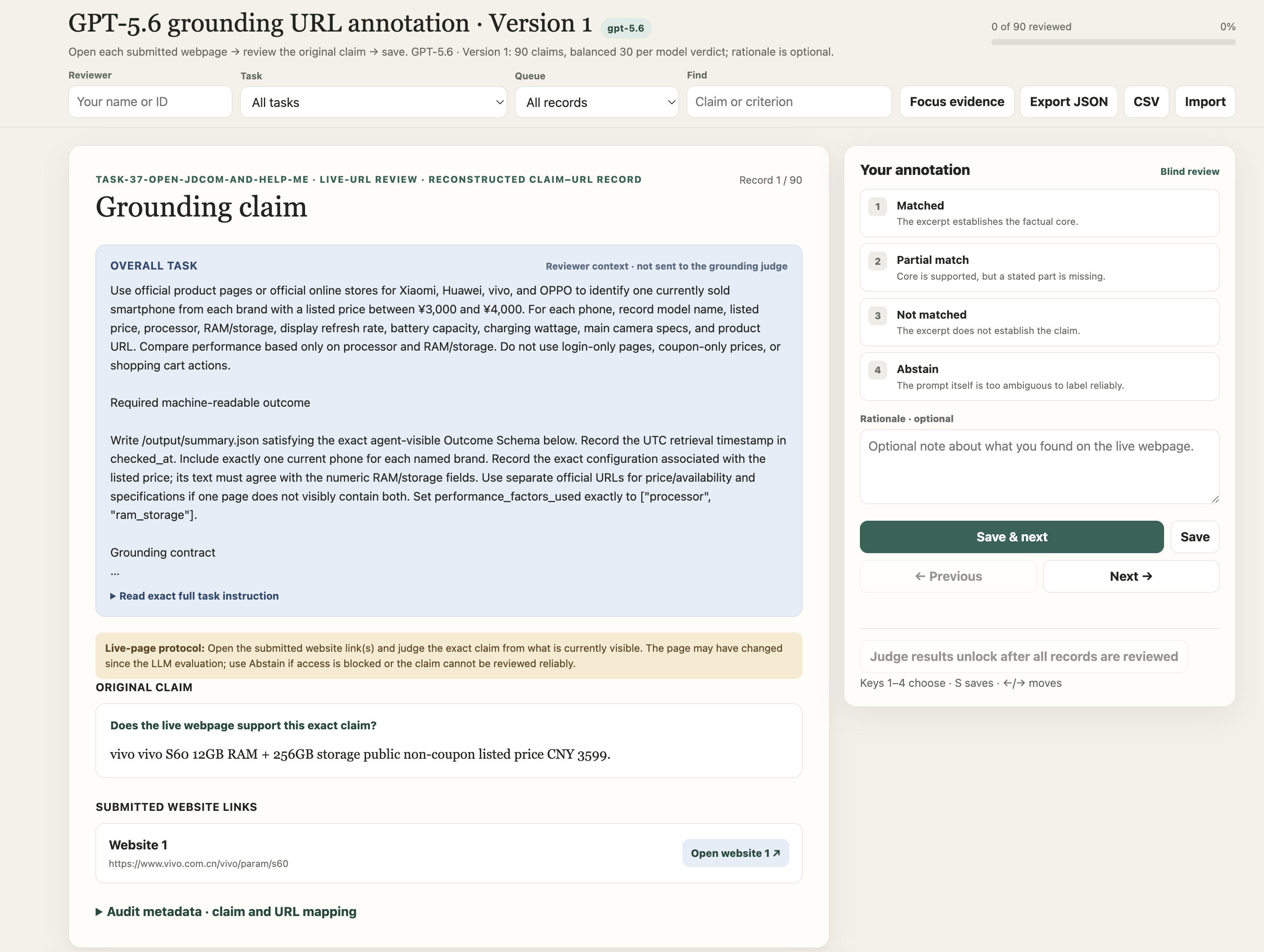}
    \caption{Human validation interface for grounding agents.}
    \label{fig:ground_annotation_webpage}
\end{figure}

\paragraph{Grounding agents validation.}
Using linearly weighted Cohen's $\kappa$, human-judge agreements for two annotators are 0.723 and 0.695 on GPT-5.4, and 0.695 and 0.667 on GPT-5.6 Sol, respectively; the corresponding inter-annotator $\kappa$ values are 0.810 and 0.890. The human validation website for grounding agents is shown in Figure \ref{fig:ground_annotation_webpage}.

\paragraph{Rubric judge validation.}

\begin{figure}[H]
    \centering
    \includegraphics[width=0.8\linewidth]{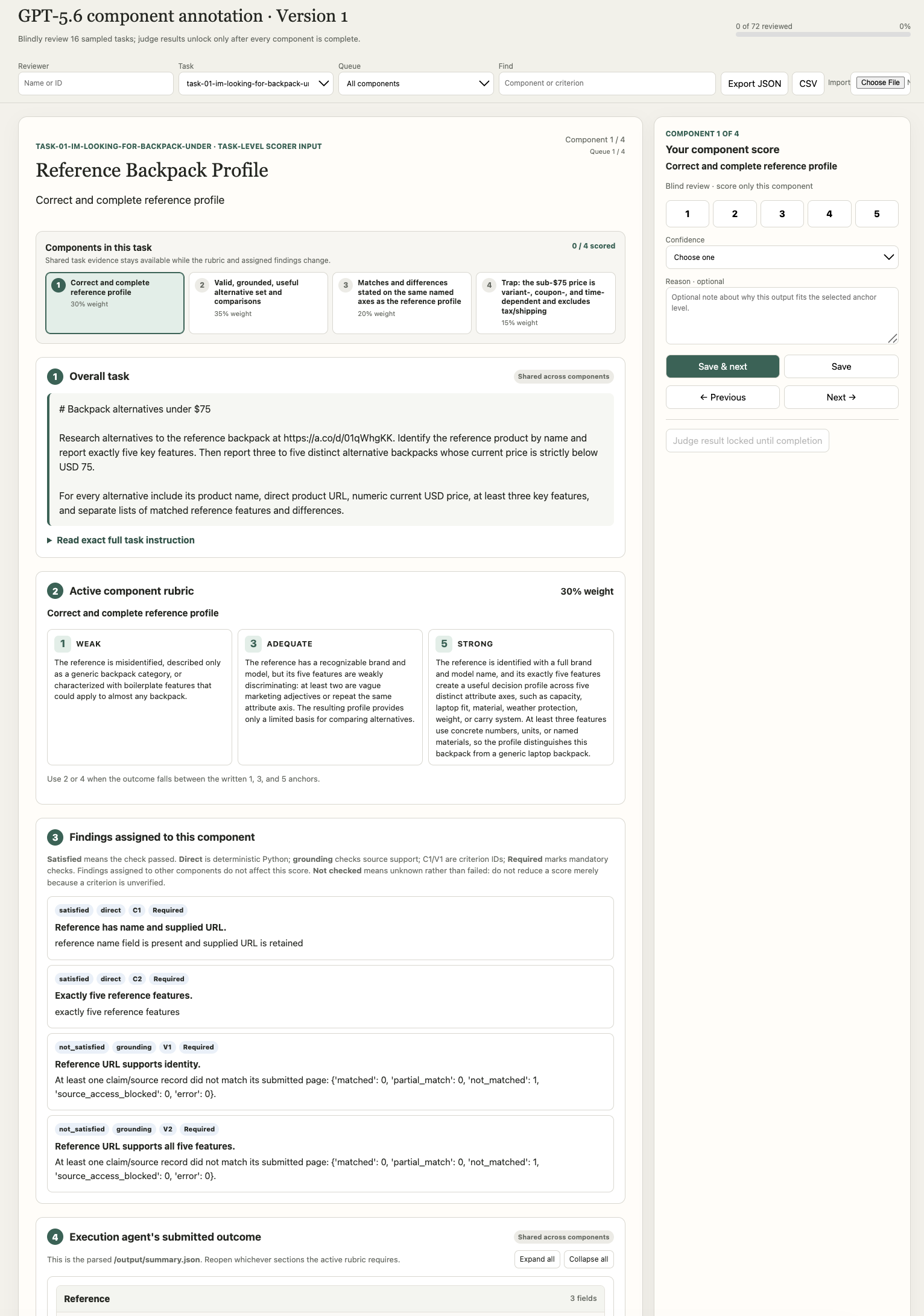}
    \caption{Human validation interface for rubric judge.}
    \label{fig:component_annotation_webpage}
\end{figure}

For GPT-5.4 outputs, the linearly weighted Cohen's $\kappa$ value between the scorer and the two annotators are 0.608 and 0.647, respectively; for GPT-5.6 Sol outputs, they are 0.625 and 0.696. The correspondent inter-annotator $\kappa$ values are 0.642 and 0.570. The human validation website for the rubric judge is shown in Figure \ref{fig:component_annotation_webpage}.

\input{sections/failure_mode_appendix_v2}

%% file: sections/appendix_threshold_sensitivity_v2.tex
\subsection{Pass-Threshold Sensitivity}
\label{sec:threshold-sensitivity}

A task passes when its overall score is at least 85\% and no rubric component scores below 3. Because component scores run from 1 to 5, an overall score of 85\% corresponds to a weighted component score of 4.25: on average, the deliverable must be better than ``good'' and reach the top anchor on a substantial share of the rubric weight. The component floor prevents strong components from masking a deficient one. To test how much our conclusions depend on these choices, we recompute pass rates 
from the 1,463 runs of the 19 primary configurations in the leaderboard
(Table~\ref{tab:agent-performance}), varying the overall-score cutoff $\tau$ from 75\% to 95\% in steps of 5 with the floor fixed at 3, and varying the floor with $\tau$ fixed at 85\%. 

We report two rank correlations over the 19 configurations of the primary leaderboard. $\rho_{\text{score}}$ is the Spearman correlation between pass rate at $\tau$ and mean overall score, which needs no cutoff. $\rho_{85}$ is the Spearman correlation between pass rate at $\tau$ and pass rate at the primary 85\% cutoff, and measures whether moving the cutoff changes the ranking we report.

Table~\ref{tab:threshold-sensitivity} shows that absolute pass rates depend strongly on the cutoff, falling from 53.86\% of all runs at $\tau=75\%$ to 31.72\% at 85\% and 7.38\% at 95\%. The ranking is much less sensitive. Between 75\% and 90\%, $\rho_{85}$ stays at or above 0.94, and Claude Opus 5.5 or Claude Fable 5.1 has the highest pass rate at every cutoff; at 85\% the two are tied at 54.55\%. Only at 95\%, where 7.38\% of runs pass and two configurations pass no task, does the ranking become noticeably less stable ($\rho_{85}=0.868$). Pass rate and mean score agree closely but not perfectly ($\rho_{\text{score}}=0.902$ at 85\%), because the pass metric rewards complete outcomes rather than partial credit: Qwen Code with GPT-6 Sol ranks seventeenth by mean score but eighth by pass rate, while Codex with GPT-6 Luna ranks eighth by mean score and tenth by pass rate. In the GPT-6 Sol reasoning-effort sweep, pass rate at 85\% rises from 29.87\% at low effort to 45.45\% at max, although neighboring levels are not strictly ordered.

The component floor changes little at the primary cutoff. A run that reaches 85\% overall rarely has a component below 3: removing the floor raises the pass rate over all runs from 31.72\% to 32.06\% (5 runs), and raising it to 4 lowers it to 28.78\%, with $\rho_{85} \geq 0.995$ in every case. At $\tau=85\%$ the floor therefore acts as a safeguard against isolated failures rather than as a second filter.

\begin{table}[t]
\centering
\caption{Pass-rate sensitivity to the overall-score cutoff $\tau$, with every component required to score at least 3. Entries are percentages over each configuration's 77 tasks. $\rho_{\text{score}}$ is the Spearman correlation with mean overall score and $\rho_{85}$ the Spearman correlation with pass rate at $\tau=85\%$, both over the 19 configurations of the primary leaderboard. Bold marks the primary cutoff.}
\label{tab:threshold-sensitivity}
\scriptsize
\setlength{\tabcolsep}{3.5pt}
\begin{tabular}{@{}lllccccc@{}}
\toprule
\textbf{Agent} &
\textbf{Model} &
\textbf{Effort} &
\multicolumn{5}{c}{\textbf{Overall-score cutoff $\tau$}} \\
\cmidrule(lr){4-8}
 & & & 75\% & 80\% & \textbf{85\%} & 90\% & 95\% \\
\midrule
\multicolumn{8}{@{}l}{\textit{Primary leaderboard}} \\
Claude Code & Claude Opus 5.5   & High   & 85.71 & 75.32 & \textbf{54.55} & 33.77 & 11.69 \\
Claude Code & Claude Fable 5.1  & High   & 81.82 & 74.03 & \textbf{54.55} & 33.77 & 15.58 \\
Codex       & GPT-6.1 Sol       & High   & 79.22 & 68.83 & \textbf{49.35} & 29.87 & 12.99 \\
Codex       & GPT-5.6 Sol       & High   & 62.34 & 50.65 & \textbf{42.86} & 29.87 & 11.69 \\
Codex       & GPT-6 Astra       & High   & 74.03 & 67.53 & \textbf{49.35} & 33.77 & 10.39 \\
Codex       & GPT-6 Sol         & High   & 62.34 & 58.44 & \textbf{38.96} & 19.48 & 6.49 \\
Claude Code & Claude Sonnet 5.5 & High   & 63.64 & 53.25 & \textbf{48.05} & 28.57 & 11.69 \\
Codex       & GPT-6 Luna        & High   & 62.34 & 46.75 & \textbf{29.87} & 14.29 & 5.19 \\
Codex       & GPT-5.6 Luna      & High   & 54.55 & 44.16 & \textbf{32.47} & 11.69 & 5.19 \\
Codex       & GPT-5.6 Terra     & High   & 48.05 & 37.66 & \textbf{29.87} & 15.58 & 9.09 \\
Codex       & GPT-5.5           & High   & 38.96 & 28.57 & \textbf{22.08} & 9.09  & 1.30 \\
Pi          & GPT-5.6 Luna      & Medium     & 38.96 & 29.87 & \textbf{20.78} & 12.99 & 6.49 \\
Claude Code & Claude Haiku 5.5  & High   & 51.95 & 40.26 & \textbf{24.68} & 15.58 & 7.79 \\
Goose       & GPT-5.6 Luna      & Medium     & 38.96 & 31.17 & \textbf{22.08} & 12.99 & 5.19 \\
Qwen Code   & GPT-5.6 Luna      & Medium     & 37.66 & 23.38 & \textbf{12.99} & 9.09  & 5.19 \\
Codex       & GPT-5.5           & Medium & 38.96 & 27.27 & \textbf{18.18} & 10.39 & 0.00 \\
Qwen Code   & GPT-6 Sol         & Medium     & 51.95 & 44.16 & \textbf{32.47} & 20.78 & 10.39 \\
Codex       & GPT-5.5           & Low    & 22.08 & 14.29 & \textbf{10.39} & 3.90  & 0.00 \\
Qwen Code   & Qwen3.7-Plus      & --     & 29.87 & 22.08 & \textbf{9.09}  & 3.90  & 3.90 \\
\midrule
\multicolumn{3}{@{}l}{All runs (19 configurations)} & 53.86 & 44.09 & \textbf{31.72} & 18.39 & 7.38 \\
\multicolumn{3}{@{}l}{$\rho_{\text{score}}$} & 0.916 & 0.909 & \textbf{0.902} & 0.823 & 0.753 \\
\multicolumn{3}{@{}l}{$\rho_{85}$}           & 0.974 & 0.979 & \textbf{1.000} & 0.942 & 0.868 \\
\midrule
\multicolumn{8}{@{}l}{\textit{Controlled Codex/GPT-6 Sol reasoning-effort sweep}} \\
Codex       & GPT-6 Sol         & Low    & 58.44 & 45.45 & \textbf{29.87} & 16.88 & 9.09 \\
Codex       & GPT-6 Sol         & Medium & 71.43 & 55.84 & \textbf{28.57} & 15.58 & 7.79 \\
Codex       & GPT-6 Sol         & High   & 62.34 & 58.44 & \textbf{38.96} & 19.48 & 6.49 \\
Codex       & GPT-6 Sol         & XHigh  & 71.43 & 58.44 & \textbf{36.36} & 18.18 & 9.09 \\
Codex       & GPT-6 Sol         & Max    & 74.03 & 63.64 & \textbf{45.45} & 24.68 & 7.79 \\
\bottomrule
\end{tabular}
\end{table}

%% file: sections/failure_mode_appendix_v2.tex
\section{Failure Mode Analysis Details}
\label{app:failure-modes}

This appendix gives the details behind the failure mode analysis in Section~\ref{sec:failure-modes}. Appendix~\ref{app:fm-unit} defines the unit of analysis, Appendix~\ref{app:fm-types} the failure types, Appendix~\ref{app:fm-judge} how the failure types were defined and how the failure-mode judge assigns them, Appendix~\ref{app:fm-prompts} the judge's prompts, Appendix~\ref{app:fm-results} the full results, and Appendix~\ref{app:fm-examples} further examples.

\subsection{Unit of Analysis}
\label{app:fm-unit}

We analyze the same runs as Table~\ref{tab:agent-performance}: one run for each configuration and task. Each task's rubric has several components (4.3 on average), and the judge scores each component from 1 to 5 with a short reason. A deduction is a component that scores below 5. When an output violates the required format, the run scores 0 and its components are not scored; we count each of its components as one deduction of the failure type \emph{output violates the required format}. Table~\ref{tab:fm-counts} gives the counts. Passing runs are included.

\begin{table}[h]
\centering\small
\setlength{\tabcolsep}{3pt}
\begin{tabular}{lrrrrrr}
\toprule
Group & Configurations & Runs & Components & Runs violating format & Deductions & Share that are a 4 \\
\midrule
Frontier & 4 & 308 & 1,320 & 1 & 756 (57.3 per 100) & 72\% \\
Low-cost & 6 & 462 & 1,980 & 33 & 1,508 (76.2 per 100) & 44\% \\
\bottomrule
\end{tabular}
\caption{Runs, components and deductions in the failure mode analysis. A deduction is a component that scores below 5; every component of a run that violates the required format counts as one deduction.}
\label{tab:fm-counts}
\end{table}




\paragraph{Rates and intervals.} We report deductions per 100 components, which removes the effect of the number of components per task. For the difference between the groups, we resample the 77 tasks with replacement 2,000 times, recompute the rate of each group on each resample, and take the 2.5th and 97.5th percentiles of the low-cost rate minus the frontier rate. In the main text we call a failure type more frequent in one group only when this interval excludes zero.

\subsection{Failure Types}
\label{app:fm-types}

Each deduction receives exactly one failure type, according to what the output is missing or gets wrong (Table~\ref{tab:fm-types}). The failure types belong to three categories that follow the three things a task requires: working out what the user needs beyond what the request states (intent recognition), finding the required information (research), and writing what was found into the final files (delivery).

\begin{table}[h]
\centering\small
\begin{tabularx}{\textwidth}{p{0.15\textwidth}p{0.22\textwidth}X}
\toprule
Category & Failure type & Definition \\
\midrule
Intent recognition & Does not look into the need & A qualification that the user needs in order to use the answer is missing, given only as a general remark, or not where the user will read it; the missing point appears neither in the delivered files nor in the model's own messages or tool calls. \\
 & Looks into the need, does not report it & As above, but the missing point appears in the model's own messages or tool calls, or in a delivered file in the wrong place, in weak wording, or for only some of the items. \\
\midrule
Research & Required items or data not found in full & Fewer items than the request requires; required fields or values absent, null or placeholders; a required type of source or a required data point not obtained. \\
 & Claim says more than the cited page & The cited page supports the core of the claim, and the claim adds details, scope, precision or certainty that the page does not state, or states the agent's own estimate or inference as if the page said it. \\
 & Value not found on the cited page & The cited page has nothing for the value, including a general page (home page, listing, search results) cited instead of the page that would show it; or no source is given for the value. \\
 & Value differs from the cited page & The cited page shows a different value, date, status, name or place for the same thing. \\
 & Items do not meet the stated conditions & Items of the wrong kind, outside the stated range or date, duplicates, not comparable, or all from one place when variety is needed. \\
 & Data selection not explained & The rule for selecting the items or records, which records were combined, the method of calculation, or the choice among candidate sources is not stated, or not stated fully. \\
\midrule
Delivery & Output violates the required format & A required field, column, file, unit or value type does not follow the stated format; also every component of an output that fails the format check. \\
 & Numbers, states or files do not agree & Two delivered files disagree, or the times, totals, counts or states within one file do not fit together. \\
 & Other & A plan, recommendation or analysis that is not worked out for the user's case; presentation; or nothing to assess. \\
\bottomrule
\end{tabularx}
\caption{The eleven failure types. Each deduction receives exactly one.}
\label{tab:fm-types}
\end{table}

\paragraph{Rules for close calls.}
\begin{enumerate}
\item \emph{What is missing, not how much.} Phrases such as ``not specific enough'' or ``could be more explicit'' say how much is missing. A missing qualification that the user needs is intent recognition; missing information, items, data or evidence is research; a plan or analysis that is not worked out for the user's case is delivery (other).
\item \emph{Required items only for what the request asks for.} A point that only the 5-point anchor asks for, and that tells the user how far to trust or how to use the answer, is intent recognition; for example, the runner-up and the price gap behind a recommendation.
\item \emph{An account of the search or the calculation is data selection}: which sources were tried, why the search stopped, which records were combined, how values were rounded or computed.
\item \emph{Trap components.} A component that tests a need the request does not state is intent recognition, unless the main shortfall is clearly source support, missing items, format or consistency. A trap mentioned only briefly or generically is still intent recognition.
\item \emph{Needs inferred from the user's situation} are intent recognition, for example the features a backpack must have for full-day birdwatching.
\item \emph{Placeholders and empty entries} are required items not found; claims resting on weak evidence say more than the cited page.
\item \emph{The three page types.} A value differs from the page only when the page shows a specific different value for the same thing. If the page supports part of the claim's specific content, the claim says more than the page; if it supports none of it, the value is not found on the page.
\item \emph{Looking into the need.} The missing point is what the judge says the output lacks, after removing everything the judge credits. A general remark that would fit many tasks of the kind, content the judge already credits, a related but different point, or only opening a page does not show that the model looked into the need.
\item \emph{Other is the last resort}, used only when no other type fits.
\end{enumerate}

\subsection{Defining and Assigning the Failure Types}
\label{app:fm-judge}

\paragraph{Defining the failure types.} We derived the failure types by hand. We sampled 100 deductions and, for each, read the judge's reason, the agent's output and its trace, which records the model's messages and tool calls, to see what the output was missing; from these cases we summarized the failure types in Table~\ref{tab:fm-types} and the rules for close calls.

\paragraph{The failure-mode judge.} The failure-mode judge is Claude Sonnet 5.5 run as an agent. Each call receives one of the prompts in Appendix~\ref{app:fm-prompts} and a batch of deductions, and returns one JSON line per deduction. It labels the deductions in three steps.
\begin{enumerate}
\item \emph{Category label} (all 2,116 deductions with a score). For each deduction the judge sees a summary of the task, the component's description, its 5-point and 3-point anchors, the score and the judge's reason. It returns one of eight labels and a short phrase saying what is missing: intent recognition; required items not found, value not supported by the cited page, conditions not met, data selection not explained; format, disagreement, other.
\item \emph{Kind of source problem} (the 612 deductions labeled as not supported by the cited page). The judge also sees up to five checks of the agent's cited claims: the claim the agent wrote, the check status, and what the checker saw on the cited page. It decides whether the claim says more than the page, the value is not on the page, the value differs from the page, or no source is given (14 deductions); these and the 2 deductions it leaves unclear are counted as not found on the cited page. If the main shortfall is not about source support, the judge moves the deduction to another label; 18 deductions moved, 4 of them to intent recognition.
\item \emph{Looking into the need} (all 659 intent recognition deductions). The judge has a command-line tool that lists, prints and searches the delivered files and searches the run's trace; each hit in the trace is marked as the model's own message, the arguments of a tool call the model made, or the text a tool returned. The judge first states the missing point and then gives one of four verdicts: the point is in a delivered file but in the wrong place, in weak wording, or for only some items (IN\_FILES); it is only in the model's own messages or tool calls (CONSIDERED); it is only in text a tool returned (SEEN\_ONLY); or it is nowhere (ABSENT). The first two verdicts mean that the model looks into the need but does not report it; the last two that it does not look into the need. Every IN\_FILES or CONSIDERED verdict is then read a second time under stricter rules (prompt 3b), and the second verdict replaces the first.
\end{enumerate}
Components of outputs that violate the required format are assigned to that failure type without the judge.

\paragraph{Trials and revisions.} Steps 1 and 3 were first run on ten trial tasks (302 deductions in step 1 and 136 intent recognition deductions in step 3). We compared the labels with our own reading of the cases and revised the prompts. After the trial batch of step 1 we added rules 4 and 5 of prompt 1, which separate required items from qualifications and from accounts of the search. After labeling all deductions, we found that the judge used \emph{other} as a catch-all; we added rules 6 to 8 of prompt 1 and relabelled the 376 deductions labeled \emph{other} with the revised prompt, which moved 111 of them to intent recognition. The stricter second reading in step 3 was added because the first reading accepted loose matches, such as a general reminder in place of the specific point; it changed 81 of the 313 verdicts it re-read. In all, 58 labels (2.7\% of the 2,116 deductions) were corrected individually after review.

\paragraph{Agreement with human labels.} On 100 deductions labeled independently by the authors, the failure-mode judge's final failure type agrees with the human label in 92\% of cases.

\subsection{Prompts of the Failure-Mode Judge}
\label{app:fm-prompts}

The prompts below are the ones used, with local file paths removed. In prompts 3a and 3b, \texttt{trace\_tool} stands for the command-line tool described in step 3.

\begin{tcblisting}{failurelisting/prompt,title={Prompt 1: category label},title after break={Prompt 1: category label (continued)}}
# Step 1: what is missing in a deduction

Each item is one rubric component that an agent's output scored below 5. You get the task summary, the component's
description, the 5-point and 3-point anchors, the score, and the judge's reason. Decide **what the judge says is missing
or wrong**, and give exactly one label.

## How to read the reason

1. Find the shortfall the judge docks the score for. It usually follows "falls short because", "but", "however",
   "does not", "misses a 5 because". Praise ("exceeds the 3-anchor", "is concrete") is not the shortfall.
2. If the reason names two shortfalls, use the one the judge gives as the main reason for not reaching the higher anchor;
   if they are equal, use the first one stated.
3. **Classify by the content of what is missing, never by how much is missing.** Phrases such as "not deep enough",
   "not specific enough", "only general", "could be more explicit" say how much is missing. Ask: missing *what*?
   - a qualification the user needs to use the answer -> `intent`
   - information, items, data or evidence -> one of the research labels
   - plan or analysis worked out for the user's case -> `other`

4. **`missing` is only for what the task request itself asks for** (items, fields, data, source types). A point that
   only the 5-point anchor asks for, and that tells the user how far to trust or how to use the answer, is `intent`:
   e.g. the runner-up and the price gap behind a recommendation, whether unreachable sources could change the ranking,
   the likely direction of bias in a fallback estimate.
5. **An account of the search or the calculation is `selection`**: which sources were tried, which could not be
   reached, why the search stopped, which records were combined, how values were rounded or computed.

6. **Trap components.** If the component description starts with "Trap:", the label is `intent` unless the judge's
   main shortfall is clearly about source support (`source`), missing items (`missing`), or a failed format or
   consistency check (`format`, `disagree`). A trap that is mentioned only briefly, generically, or without acting on it
   in the plan is still `intent`.
7. **Needs inferred from the user's situation are `intent`.** When the component checks whether the output works out
   what the user needs from their stated situation (e.g. the features a backpack must have for full-day birdwatching,
   eligibility routes for a postdoc applicant), a shortfall in that inference is `intent`.
8. **`other` is the last resort.** Use it only when none of the content labels fits: a plan, recommendation or
   analysis that is generic or not worked out, presentation, or nothing to assess. Placeholders or empty entries are
   `missing`; claims resting on weak or snippet-level evidence are `source`.

## Labels

**`intent`  -  a qualification the user needs is missing, only general, or not where the user will read it.**
The answer itself is there, but something the user must know in order to use it correctly is not said, is said only as
a generic remark, or sits where the user will not see it (for example in a notes field and not in the table). Typical
content: what a number covers (combined groups, scope, definition, units of a rate); that a status or price may change
or is dated; limits of the evidence (bias, small numbers, unverified, assumption, estimate); preconditions (laws, local
rules, eligibility, licences, safety, age, access); costs or factors not counted (tax, shipping, fees); a known risk
that the plan does not act on. A component whose description starts with "Trap:" tests such a need: if the shortfall is
about that trap, the label is `intent`.
- e.g. "does not identify the smallest denominator among the five employers or explain the fragility of that rate" -> intent
- e.g. "does not say that the Software Developers figure also covers QA analysts and testers" -> intent
- e.g. "does not discuss whether shipping could change the ranking across listings" -> intent

**`missing`  -  required items or data are not found in full.**
Fewer items than required, required fields or values absent or null, a required type of source absent, a required
data point not obtained, the search covered too little to meet the stated requirement.
- e.g. "only eight items, all arXiv; the task asks for at least three source types" -> missing
- e.g. "no usable stay-specific nightly price for the listings" -> missing

**`source`  -  a stated value or claim is not supported by the cited page.**
The cited page does not show the value, shows a different value, supports only part of the claim, or is a generic page
(home page, search page) rather than the page with the value; claims or labels rest on inference rather than evidence.
- e.g. "reports the primary completion date as the study completion date" -> source
- e.g. "cost and difficulty labels rely on inference or broad claims rather than evidenced details" -> source

**`conditions`  -  the delivered items do not meet the stated conditions.**
Items are of the wrong kind, outside the stated range or date, duplicates, not comparable, all from one place when
variety is needed, or otherwise unsuitable for what the user asked.
- e.g. "all five offers are from the same retailer" -> conditions

**`selection`  -  how the data or items were selected or computed is not explained.**
The selection rule, which records were counted or combined, the method of calculation, or why one candidate source was
chosen is not stated or not stated fully.
- e.g. "does not identify the exact Employer Data Hub record selection rule for each company" -> selection
- e.g. "the selection account reviews only one annual candidate source" -> selection

**`disagree`  -  numbers, states or files do not agree.**
Two delivered files disagree; within one file, times, totals, counts or states do not fit together (overlapping time
windows, sums that do not add up, a status contradicting another field); a stated value conflicts with the agent's own
calculation; a fixed consistency check fails.
- e.g. "the JSON names one speaker and the document is built around a different person" -> disagree

**`format`  -  the output violates the required format.**
A required field, column, file, unit or value type does not follow the format the task states, as reported by the judge.

**`other`  -  a plan or analysis that is not worked out for the user's case, presentation, or nothing to assess.**
Recommendations, plans or analysis that would fit many other cases, are not specific or actionable enough, or are not
sequenced into concrete steps; readability or presentation; the situation the component checks did not occur.
- e.g. "the activities remain adaptable to many other photography exhibitions" -> other
- e.g. "some recommendations remain generic rather than sequenced into a launch plan" -> other

## Output

One JSON object per line, in the order of the batch:

```
{"did": "D0123", "label": "intent", "missing": "smallest-denominator caveat for one employer"}
```

`missing` is a short phrase (at most 15 words) saying what is missing or wrong. Use only the eight labels above.
\end{tcblisting}

\begin{tcblisting}{failurelisting/prompt,title={Prompt 2: kind of source problem},title after break={Prompt 2: kind of source problem (continued)}}
# Step 2: kinds of "value not supported by the cited page"

Each item is one rubric component that an agent's output scored below 5, and that a first pass labelled as
"a stated value or claim is not supported by the cited page". You get the task summary, the component description, the
5-point anchor, the score, the judge's reason, the first pass's short note, and up to five claim checks from the
source checker (`records_shown`: the claim the agent wrote, the checker's status, and what the checker saw on the cited
page). `record_counts` gives all checks for the component. If `checks_scope` says "whole run", the checks are for the
whole run and may not all concern this component: rely more on the judge's reason.

Give exactly one label: one of four kinds of source problem, or a move to another category when the main shortfall is
not about source support at all.

## Source kinds

**`contradicts`  -  the written value differs from the page.**
The cited page shows a different value, date, status, name, place or item for the same thing.
- page shows write operations at $0.0175, the agent wrote $0.065 -> contradicts
- page shows a prerequisite rule, the agent wrote that there are no prerequisites -> contradicts
- page is an Annapolis news digest, the agent wrote it as an Austin angle -> contradicts

**`not_on_page`  -  the cited page does not show the value.**
The cited page has nothing for the value or claim: either the right kind of page that simply lacks it, or a general
page (home page, all-jobs listing, search results, lab front page, archive file) cited instead of the page that would
show it.
- product page readable, but no base storage of 512GB shown -> not_on_page
- lab home page cited for a named student and research focus -> not_on_page

**`overclaim`  -  the claim says more than the page.**
The page supports the core of the claim, but the written claim adds details, scope, precision or certainty that the
page does not state. Also: the agent's own estimate, inference, label or recommendation written as if the page said it.
- page confirms a 7-year warranty, the agent adds a breakdown of 5 years for tops and 1 year for grommets -> overclaim
- page supports strong damage, the agent adds an anti-heal strength -> overclaim
- cost or difficulty labels that rest on the agent's inference rather than on the page -> overclaim

**`no_source`  -  no source is given for some values.**
Values are given without any source, the output says they came from a source it did not cite, or an estimate is given
with no stated source or basis.
- heights taken from spec databases while the cited pages are different ones -> no_source
- travel times for some legs explicitly unsourced -> no_source
- estimated study hours with no source or workload basis -> no_source

## How to decide

1. Read the judge's reason: which claims does it say are not supported, and how? That decides the label when it is
   clear.
2. If the reason only says "grounding shows partial / not matched support", use the claim checks:
   `partial_match` usually means `overclaim`; `not_matched` means `contradicts` when the checker saw a different specific
   value for the same thing, otherwise `not_on_page`.
3. If the checks are mixed, choose the kind that covers the claims the judge treats as central; if the judge names none,
   choose the kind most common among the `not_matched` and `partial_match` checks.
4. Ignore `page_not_read` checks: they are not evidence either way.
5. `contradicts` needs a specific different value seen on the page. "Does not show X" alone is `not_on_page`.
6. `not_on_page` vs `overclaim`: if the page supports part of the claim's specific content, `overclaim`; if it supports
   none of it, `not_on_page`.

## Move to another category (use sparingly)

Use `move:<label>` only when the judge's **main** shortfall for this component is not whether written values are
supported by sources. If source support is the main shortfall, keep a source kind even if other issues are mentioned.
- `move:intent`  -  a qualification the user needs is not given (what a number covers, that it may change, limits of the
  evidence, preconditions, costs not counted, a trap not acted on).
- `move:missing`  -  required items, fields or values are absent or null; too few items.
- `move:conditions`  -  items do not meet the stated conditions (wrong kind, out of range, duplicates).
- `move:selection`  -  the account of how sources or records were chosen, combined or computed is missing or incomplete.
- `move:disagree`  -  numbers, states or files within the output do not agree with each other.
- `move:other`  -  a plan, recommendation or analysis not worked out for the user's case; presentation.

## Unclear

`unclear`  -  the only problem named is that pages could not be read or checked, or the reason does not say what is
unsupported and there are no usable checks.

## Output

One JSON object per line, in the order of the batch:

```
{"did": "D0123", "label": "not_on_page", "note": "base storage not shown on Apple newsroom page"}
```

`note` is a short phrase (at most 15 words) saying which value or claim, and what is wrong with it.
Labels: contradicts, not_on_page, overclaim, no_source, move:intent, move:missing, move:conditions, move:selection,
move:disagree, move:other, unclear.
\end{tcblisting}

\begin{tcblisting}{failurelisting/prompt,title={Prompt 3a: looking into the need, first reading},title after break={Prompt 3a: looking into the need, first reading (continued)}}
# Step 3a: did the model look into the need? (first reading)

## Context

A benchmark judge scored components of an agent's output and wrote a reason for each score below 5. Every deduction in your worklist is about a qualification, limitation, basis or reminder that the user needs and that the deliverable does not give properly. We want to know, from the record of the run, whether the model ever had this point in view.

Each entry in your worklist looks like this:

```
### D0123
task: task-115-... | config: <configuration> | component: combined_occupation_group_figures | score 1
component checks: ...
5-point description: ...
judge's reason: ...
```

## Tool

Use the trace tool with the task name and the configuration exactly as given in the entry:

```
trace_tool file TASK CONFIG                      # list the delivered files
trace_tool file TASK CONFIG NAME 6000            # print a delivered file
trace_tool out  TASK CONFIG 'REGEX' 150 8        # regex hits in the delivered files
trace_tool fx   TASK CONFIG 'REGEX' 150 12       # regex hits in the run's trace
```

`fx` labels each hit: `own` is a message the model wrote; `call` is the arguments of a tool call the model made (search queries, commands, code); `result` is what a tool returned (page text, file listings). The regex is case-insensitive.

## Steps for each entry

1. From the judge's reason and the 5-point description, write the **specific point** that is missing, in one short phrase. Example: "the Software Developers figure also covers QA analysts and testers".
2. Search the delivered files for it (`out`).
3. Search the trace for it (`fx`). Use a regex with two to four alternatives, including synonyms. If the first regex finds nothing, try one more with different words.
4. Choose one verdict.

## Verdicts

**IN_FILES**  -  The specific point is stated in at least one delivered file, though not in the place or not as clearly as the rubric wants (for example in the JSON and not in the table, or only in a notes field).

**CONSIDERED**  -  The specific point is not in the delivered files, and an `own` or `call` hit shows that the model had it in mind: it searched for it, ran a command about it, or wrote about it in a message.

**SEEN_ONLY**  -  The specific point appears only in `result` text that a tool returned. The model itself never refers to it.

**ABSENT**  -  Nothing in the delivered files or the trace touches the specific point.

## Rules

- The point must be the specific one. A general remark does not count. Example: for a missing explanation of a city's owner-occupancy rule for short-term rentals, "check that the rental is legal" is general and does not count; a search for the city's short-term-rental ordinance does.
- Ignore hits in tool boilerplate: file listings, plugin paths, system text.
- A URL or page title is not enough by itself. The words must show the point.
- If the delivered files state the point, use IN_FILES even when the trace also shows it.
- Do not judge whether the judge is right. Only establish where the point appears.

## Output

One JSON object per entry, one per line, in the order of the worklist:

```
{"id":"D0123","point":"the Software Developers figure also covers QA analysts and testers","verdict":"CONSIDERED","evidence":"step 5 call: search query 'software developers quality assurance analysts testers'"}
```

- `evidence`: at most 30 words. Name the step and the kind of hit, or the file. For ABSENT, list the regexes you tried.

## Further rules

- Each entry also has a `task summary` and an `earlier note on what is missing` from step 1. The note is a hint only; the judge's reason decides the specific point.
- **Same point, stated incompletely.** If a delivered file states the specific missing point but only for some of the items, only loosely, only in passing, or only in another file or field, that is IN_FILES.
- **A different point does not count.** If the file states something the judge already credits, and the judge says a different point is missing, search for the different point.
- **A general remark does not count.** A remark that would fit any task of this kind ("verify before booking", "check local rules", "prices may change") does not show the specific point, unless the specific point is exactly that remark applied to the item in question.
- If the judge names several missing points, any one of them found in files or in `own`/`call` text is enough for IN_FILES or CONSIDERED.
- Several entries for the same task and component usually miss the same point. Write the point the same way for each, but search each run's own files and trace.
- If a run has no delivered files, say so in `evidence` and decide from the trace.
\end{tcblisting}

\begin{tcblisting}{failurelisting/prompt,title={Prompt 3b: looking into the need, second reading},title after break={Prompt 3b: looking into the need, second reading (continued)}}
# Step 3b: is the "looked into" verdict right? (second reading)

A first reader coded each intent-recognition deduction with one of four verdicts (see the step 3a prompt): IN_FILES, CONSIDERED, SEEN_ONLY, ABSENT. You re-check the entries the first reader coded IN_FILES or
CONSIDERED, because the first readers often accepted loose matches. Use the same trace tool as in step 3a, with the full task name and configuration.

Each entry in your worklist gives the task, config, component, score, the 5-point description, the judge's reason,
and the first reader's point, verdict and evidence.

## Step 1. Fix the missing point

From the judge's reason, write the **missing point**: what the judge says the output lacks, **after removing everything
the judge credits**. Judges usually write "it does X ... but falls short because it does not Y". The missing point is
Y, not X. If the first reader's point describes X or the topic in general, rewrite it.

## Step 2. Read the actual wording in the delivered files

Open the delivered files (`trace_tool out` with a regex for the missing point, and `trace_tool file` when needed) and find the
closest sentence or field. Quote it in `evidence`.

Keep **IN_FILES** only if that wording states the missing point Y itself, and the problem is only one of these:
- it is in the wrong place (another file, a JSON field, a notes section, a footnote) rather than where the rubric wants it;
- it is worded weakly or indirectly, but a reader would still get Y from it;
- it is given for some of the items but not all of them.

It is **not** IN_FILES when the wording is:
- what the judge already credits (X), or the topic without Y;
- a general remark that would fit many tasks of this kind ("local taxes may apply", "verify before booking", "pay
  varies by arrangement", "check visibility", "prices may change", "confirm with your institution"), unless Y is
  exactly that remark;
- a related but different point (for example, stating the China-origin assumption when Y is the rate for imports
  that are not of Chinese origin; giving the costs of alternatives when Y is the condition under which to switch).

## Step 3. If not IN_FILES, search the trace for Y

Use `trace_tool fx` with two to four alternatives for Y. Then choose:
- **CONSIDERED**: an `own` message (including the model's final chat message) or a `call` (search query, command,
  code, fetch prompt) shows the model had Y itself in mind; the same strictness applies (a general remark or X does
  not count). Opening a page or fetching a URL is not enough by itself: the words of the query, prompt, code or
  message must name Y. Extracting a raw field (for example a "YTD return" field) does not show Y unless Y is about that
  field's basis and the model says so;
- **SEEN_ONLY**: Y appears only in `result` text from a tool;
- **ABSENT**: Y appears nowhere.

## Output

One JSON object per entry, in worklist order, to your output file:

```
{"id":"D0123","point":"<missing point Y>","first":"IN_FILES","verdict":"ABSENT","evidence":"closest file wording: '...' (general remark); fx regexes a|b|c: no own/call hit"}
```

`evidence` at most 40 words. Quote the closest file wording whatever the verdict.
\end{tcblisting}

\subsection{Full Results}
\label{app:fm-results}

Table~\ref{tab:fm-results} gives every failure type with its interval and its share of deductions that are a 4, and Table~\ref{tab:fm-config} gives the rates for each configuration. Two comparisons within the same harness or model support the main finding on intent recognition. Within Codex, GPT-6 Luna has 17.0 deductions per 100 components for needs that it does not look into, against 11.8 for GPT-6 Sol at max effort and 7.3 for GPT-6.1 Sol, while its rate for needs that it looks into but does not report (8.5) is close to theirs (8.8 and 7.3). GPT-5.6 Luna has between 10.3 and 15.8 deductions per 100 components for needs that it does not look into in the three harnesses where it runs. Claude Haiku 5.5 is the exception in the low-cost group: its rate for needs that it does not look into (6.4) is close to those of Claude Opus 5.5 and Claude Fable 5.1 (4.2 and 6.7), and most of its extra deductions come from the 9 of its 77 runs whose output violates the required format (13.9 deductions per 100 components).

\begin{table}[h]
\centering\footnotesize
\setlength{\tabcolsep}{3pt}
\renewcommand{\arraystretch}{1.08}
\begin{tabularx}{\textwidth}{@{}>{\raggedright\arraybackslash}Xrrrrrr@{}}
\toprule
 & \multicolumn{2}{c}{Per 100 components} & \makecell[c]{Low-cost $-$\\frontier} & \multicolumn{2}{c}{Share that are a 4} & \makecell[c]{Deductions\\F / L} \\
\cmidrule(lr){2-3}\cmidrule(lr){5-6}
Failure type & Frontier & Low-cost & \makecell[c]{95\%\\interval} & Frontier & Low-cost &  \\
\midrule
\textbf{Intent recognition} & 16.5 & 22.3 & +3.5 to +7.9 & 69\% & 50\% & 218 / 441 \\
\quad Does not look into the need & 7.5 & 12.6 & +3.4 to +6.9 & 66\% & 42\% & 99 / 250 \\
\quad Looks into the need, does not report it & 9.0 & 9.6 & -1.3 to +2.6 & 71\% & 60\% & 119 / 191 \\
\textbf{Research} & 29.5 & 32.5 & +0.5 to +5.6 & 75\% & 46\% & 389 / 643 \\
\quad Required items or data not found in full & 3.5 & 7.3 & +2.1 to +5.7 & 59\% & 23\% & 46 / 145 \\
\quad Claim says more than the cited page & 9.1 & 5.4 & -5.5 to -2.0 & 88\% & 80\% & 120 / 106 \\
\quad Value not found on the cited page & 6.4 & 6.5 & -1.4 to +1.4 & 61\% & 33\% & 85 / 128 \\
\quad Value differs from the cited page & 3.9 & 5.2 & -0.5 to +3.1 & 79\% & 41\% & 52 / 103 \\
\quad Items do not meet the stated conditions & 4.1 & 5.2 & +0.0 to +2.1 & 76\% & 54\% & 54 / 102 \\
\quad Data selection not explained & 2.4 & 3.0 & -0.5 to +1.6 & 81\% & 61\% & 32 / 59 \\
\textbf{Delivery} & 11.3 & 21.4 & +6.6 to +13.8 & 70\% & 34\% & 149 / 424 \\
\quad Output violates the required format & 1.1 & 8.2 & +4.1 to +10.4 & 57\% & 4\% & 14 / 163 \\
\quad Numbers, states or files do not agree & 3.3 & 4.1 & -0.5 to +2.1 & 47\% & 29\% & 43 / 82 \\
\quad Other & 7.0 & 9.0 & +0.3 to +4.3 & 84\% & 64\% & 92 / 179 \\
\bottomrule
\end{tabularx}
\caption{Deductions by failure type over the 77 tasks. The interval is the 95\% bootstrap interval over tasks of the low-cost rate minus the frontier rate. Components of outputs that violate the required format score 0 and count among the deductions that are not a 4.}
\label{tab:fm-results}
\end{table}

\begin{sidewaystable}[p]
\centering\footnotesize
\setlength{\tabcolsep}{4pt}
\renewcommand{\arraystretch}{1.12}
\begin{tabular}{@{}lrrrrrrrrrrrr@{}}
\toprule
 & & \multicolumn{2}{c}{Intent recognition} & \multicolumn{6}{c}{Research} & \multicolumn{3}{c}{Delivery} \\
\cmidrule(lr){3-4}\cmidrule(lr){5-10}\cmidrule(lr){11-13}
Configuration & All & \makecell[c]{Not\\looked} & \makecell[c]{Not\\reported} & \makecell[c]{Not\\found} & \makecell[c]{Says\\more} & \makecell[c]{Not on\\page} & Differs & Conditions & Selection & Format & Disagree & Other \\
\midrule
Claude Code / Opus 5.5 & 53.6 & 4.2 & 11.2 & 3.3 & 9.7 & 5.5 & 3.9 & 4.2 & 2.1 & 0.6 & 2.7 & 6.1 \\
Claude Code / Fable 5.1 & 56.1 & 6.7 & 8.8 & 1.5 & 11.2 & 6.7 & 3.9 & 4.8 & 2.7 & 1.5 & 3.6 & 4.5 \\
Codex / GPT-6.1 Sol & 57.9 & 7.3 & 7.3 & 6.4 & 7.0 & 7.0 & 4.5 & 3.9 & 0.9 & 1.8 & 3.3 & 8.5 \\
Codex / GPT-6 Sol (max) & 61.5 & 11.8 & 8.8 & 2.7 & 8.5 & 6.7 & 3.3 & 3.3 & 3.9 & 0.3 & 3.3 & 8.8 \\
\midrule
Claude Code / Haiku 5.5 & 74.8 & 6.4 & 10.9 & 5.8 & 7.0 & 8.2 & 3.9 & 4.8 & 2.7 & 13.9 & 3.6 & 7.6 \\
Codex / GPT-6 Luna & 70.0 & 17.0 & 8.5 & 3.9 & 5.5 & 5.2 & 6.1 & 4.2 & 3.6 & 1.2 & 3.6 & 11.2 \\
Goose / GPT-5.6 Luna & 75.2 & 13.9 & 9.7 & 8.5 & 4.8 & 8.5 & 4.8 & 4.8 & 3.3 & 6.7 & 2.4 & 7.6 \\
Pi / GPT-5.6 Luna & 76.7 & 15.8 & 8.5 & 6.7 & 3.3 & 7.0 & 8.8 & 4.5 & 3.6 & 4.5 & 3.6 & 10.3 \\
Qwen Code / GPT-5.6 Luna & 77.0 & 10.3 & 9.4 & 11.5 & 5.8 & 4.2 & 3.9 & 6.4 & 2.1 & 8.8 & 4.2 & 10.3 \\
Qwen Code / Qwen3.7-Plus & 83.3 & 12.4 & 10.9 & 7.6 & 5.8 & 5.8 & 3.6 & 6.1 & 2.4 & 14.2 & 7.3 & 7.3 \\
\bottomrule
\end{tabular}
\caption{Deductions per 100 components by configuration (330 components each). The first four rows are the frontier group and the last six the low-cost group. Column names abbreviate the failure types in Table~\ref{tab:fm-types}, in the same order.}
\label{tab:fm-config}
\end{sidewaystable}

\subsection{Further Examples}
\label{app:fm-examples}

The main text gives one example for most failure types. The examples below cover the remaining types and show how the rules for close calls apply.

\paragraph{Data selection not explained.} A task asks for eight portfolio companies of a venture firm, selected under a stated rule that avoids hand-picking, with comparable fields for each company and an analysis of where the portfolio concentrates. Claude Fable 5.1 delivers eight companies with complete fields, but it does not say how they were chosen, and its analysis speaks about the whole portfolio rather than this sample (score 3). The items are all present, and what is missing is how they were selected, so the type is data selection not explained rather than required items not found (rule 3).

\paragraph{Other.} A task asks for a map of boutique-fitness studios in Denver and a summary of positioning opportunities drawn from the listed competitors. GPT-6 Luna in Codex derives its summary from the neighborhoods and class types in its list and bounds it to the sample, but its opportunity statements, such as making recovery or beginner support concrete, are not tied to specific competitors (score 4). The data are complete and supported; what is missing is analysis worked out for this case, so the type is other.

\paragraph{Intent recognition rather than a required item.} A task asks for a recommendation among three gift options using only their visible price, rating, review count and category relevance. The 5-point anchor also asks for at least one disadvantage of the winner relative to a specific runner-up. Claude Opus 5.5 explains why its pick leads on rating and review count, but it does not name such a disadvantage (score 4). Only the anchor asks for this point, and it tells the user how far to trust the pick, so the type is intent recognition (rule 2). In its final message the model compares all three options, which the delivered file does not, so the trace verdict is that it looks into the need but does not report it.

\paragraph{A trap and a general remark.} The component of a short-term-rental task near Asheville, North Carolina, is a trap: the city restricts whole-home short-term rentals that are not owner-occupied, so the revenue figures may not be attainable for a new buyer. Claude Haiku 5.5 does not address this (score 1); its final message says only that the use must be allowed under the applicable city or county rules for that exact parcel, and that it did not verify them. The type is intent recognition (rule 4). The remark would fit any rental task and does not name the owner-occupancy rule, and nothing in the files or the trace does, so the model does not look into the need (rule 8).